\documentclass{article}

\usepackage{amsmath}
\usepackage{amsfonts}
\usepackage{amssymb}
\usepackage{amsthm}
\usepackage{xcolor}

\usepackage{microtype,amssymb,amsmath}
\usepackage{graphicx}
\usepackage{subfigure}
\usepackage{booktabs} 
\newcommand{\cameraReady}[1]{}
\newcommand{\removed}[1]{}
\newcommand{\added}[1]{#1}
\usepackage{hyperref}

\usepackage[accepted]{icml2019}
\usepackage{enumitem}
\usepackage{subfigure}
\usepackage{hyperref}       
\usepackage{url}            
\usepackage{booktabs}       
\usepackage{amsfonts}       
\usepackage{nicefrac}       
\usepackage{microtype}      
\usepackage{amsmath}
\usepackage{algorithm,algorithmic}
\usepackage{graphicx}
\usepackage{amsmath,amssymb,amsthm}
\usepackage{algorithm,algorithmic}
\usepackage{adjustbox}
\usepackage{graphics}
\usepackage{xcolor}
\usepackage{framed}
\usepackage{graphicx}
\graphicspath{ {images/} }
\usepackage{tikz}
\usepackage{pdfpages}
\usepackage{wrapfig}

\usepackage{mathtools}
\usepackage[acronym, nowarn]{glossaries}
\glsdisablehyper{}
\newacronym{ES}{ES}{evolution strategies}
\newacronym{RL}{RL}{reinforcement learning}
\newacronym{IP}{IP}{Integer Program}
\newacronym{CO}{CO}{combinatorial optimization}
\newacronym{MDP}{MDP}{{M}arkov decision process}
\newacronym{MC}{MC}{{M}onte {C}arlo}
\newacronym{ML}{ML}{{M}achine learning}
\newacronym{bnc}{B\&C}{{B}ranch-and-Cut}
\newacronym{bnb}{B\&B}{{B}ranch-and-Bound}
\newacronym{CP}{CP}{{C}utting plane}
\newacronym{LP}{LP}{Linear program}
\newacronym{LR}{LR}{Linear Relaxation}
\newacronym{TSP}{TSP}{{T}raveling {S}alesman {P}roblem}
\newacronym{igc}{IGC}{Integrality Gap Closure}

\usepackage{natbib}
\usepackage[T1]{fontenc}
\usepackage[utf8]{inputenc}
\usepackage{tabularx,ragged2e,booktabs,caption}
\newcolumntype{C}[1]{>{\Centering}m{#1}}

\usepackage{caption}

\usepackage{thmtools}
\usepackage{thm-restate}
\usepackage{cleveref}

\newcommand{\sacomment}[1]{}

\newcommand{\yfcomment}[1]{}

\begin{document}

\twocolumn[
\icmltitle{Reinforcement Learning for Integer Programming: Learning to Cut}



\icmlsetsymbol{equal}{*}

\begin{icmlauthorlist}
\icmlauthor{Yunhao Tang}{cu}
\icmlauthor{Shipra Agrawal}{cu}
\icmlauthor{Yuri Faenza}{cu}
\end{icmlauthorlist}

\icmlaffiliation{cu}{Columbia University, New York, USA}
\icmlcorrespondingauthor{yt2541@columbia.edu}{Yunhao}

\icmlkeywords{Machine Learning, ICML}

\vskip 0.3in
]



\printAffiliationsAndNotice{}  

\begin{abstract}
Integer programming is a general optimization framework with a wide variety of applications, e.g., in scheduling, production planning, and graph optimization. As Integer Programs (IPs) model many provably hard to solve problems, modern IP solvers  rely on heuristics. These heuristics are often human-designed, and  tuned over time using experience and data.  
The goal of this work is to show that the performance of those \removed{solvers} \added{heuristics} can be greatly enhanced using reinforcement learning (RL). In particular, we investigate a specific methodology for solving IPs, known as the cutting plane method. This method is employed as a subroutine by all modern IP solvers.  We present a deep RL formulation, network architecture, and algorithms for intelligent adaptive selection of cutting planes (aka cuts). Across a wide range of IP tasks, we show that our trained RL agent significantly outperforms human-designed heuristics. \removed{We also demonstrate  generalization properties across instance size and problem classes.} \added{Further, our experiments show that the RL agent adds meaningful cuts (e.g.~resembling cover inequalities when applied to the knapsack problem), and has generalization properties across instance sizes and problem classes.}
The trained agent is also demonstrated to  benefit the popular downstream application of cutting plane methods in Branch-and-Cut algorithm, which is the backbone of state-of-the-art commercial IP solvers.

\end{abstract}
\section{Introduction}

Integer Programming is a very versatile modeling tool for discrete and combinatorial optimization problems, with applications in scheduling and production planning, 
among others. In its most general form, an \gls{IP} minimizes a linear objective function over a set of integer points that satisfy a finite family of linear constraints. Classical results in polyhedral theory  (see e.g.~\citet{conforti2014integer}) imply that {\it any} combinatorial optimization problem with finite feasible region can be formulated as an \gls{IP}. 
 Hence, \gls{IP} is a natural model for many graph optimization problems, such as the celebrated \gls{TSP}\removed{, maximum cut, minimum vertex cover, etc}.

Due to the their generality, \gls{IP}s can be very hard to solve in theory (NP-hard) and in practice. 
There is no polynomial time algorithm with guaranteed solutions for all \gls{IP}s. 
It is therefore crucial to develop efficient heuristics for solving specific classes of \gls{IP}s. \Gls{ML}  arises as a natural tool for tuning those heuristics. Indeed, the application of \gls{ML} to discrete optimization has been a topic of significant interest in recent years, with a range of different approaches in the literature \added{\citep{bengio2018machine}}
\removed{(see \citet{bengio2018machine} for a survey).}

One set of approaches focus on directly learning the mapping from an \gls{IP}  instance to an approximate optimal solution  \citep{vinyals2015pointer,bello2017neural,nowak2017note,kool2018attention}. These methods implicitly learn a solution procedure for a problem instance as a function prediction. These approaches are attractive for their blackbox nature and wide applicability. 
At the other end of the spectrum are approaches which embed \gls{ML} agents as subroutines in a problem-specific,  human-designed algorithms \removed{for combinatorial optimization problems} \citep{dai2017learning,li2018combinatorial}. \gls{ML} is used to improve some heuristic parts of that algorithm.\removed{For example, \citet{dai2017learning} guide and tune greedy algorithms for \gls{TSP}, minimum vertex cover, and maximum cut problems using \gls{RL}.}
These approaches can benefit from algorithm design know-how for many important and difficult classes of problems, but their applicability is limited by the\removed{applicability of the} specific (e.g.~greedy) algorithmic techniques. 


In this paper, we take an approach with the potential to combine the benefits of both \removed{the} lines of research described above. We design a \gls{RL} agent to be used as a subroutine in a popular algorithmic framework for \gls{IP} called the \emph{cutting plane method}, thus building upon and benefiting from decades of research and understanding of this fundamental\removed{ algorithmic} approach for solving \gls{IP}s. The specific cutting plane algorithm that we focus on is Gomory's  method \citep{gomory1960algorithm}. \removed{In theory,} Gomory's cutting plane method \removed{can} \added{is guaranteed to}  solve {\it any} \gls{IP} in finite time, thus our approach enjoys wide applicability. In fact, we demonstrate that our trained RL agent can even be used, in an almost blackbox manner, as a subroutine  in another powerful \gls{IP} method called \gls{bnc}, to obtain significant improvements. A recent line of work closely related to our approach includes \citep{khalil2016learning,khalil2017learning,balcan2018learning}, where  {\it supervised learning} is used to improve
{\it branching heuristics} in the \gls{bnb} framework for \gls{IP}. To the best of our knowledge, no work on focusing on pure selection of (Gomory) cuts has appeared in the literature.\sacomment{please check recent literature for more related work in 2019}

 Cutting plane and \gls{bnc} methods rely on the idea that every \gls{IP} can be relaxed to a \gls{LP} by dropping the integrality constraints, and efficient algorithms for solving \gls{LP}s are available. Cutting plane methods iteratively add \emph{cuts} to the \gls{LP}s, which are linear constraints that can tighten the \gls{LP} relaxation by eliminating some part of the feasible region, while preserving the \gls{IP} optimal solution.
\gls{bnc} methods are based on combining \gls{bnb} with cutting plane methods and other heuristics; see Section \ref{sec:background} for details. 
 
 Cutting plane methods have had a tremendous impact on the development of algorithms for \gls{IP}s, e.g., these methods were employed to solve the first non-trivial instance of \gls{TSP}~\citep{dantzig1954solution}. The systematic \removed{introduction}\added{use} of cutting planes has moreover been responsible for the huge speedups of \gls{IP} solvers in the 90s~\citep{balas1993lift,bixby}. Gomory cuts and other cutting plane methods are today widely employed in modern solvers, most commonly as a subroutine of the \gls{bnc} methods that are the backbone of state-of-the-art commercial \gls{IP} solvers like Gurobi and Cplex \citep{gurobi2015gurobi}. 
  However, despite the amount of research on the subject, deciding \emph{which} cutting plane to add remains a non-trivial task. As reported in~\citep{dey2018theoretical}, ``several issues need to be considered in the selection process [...] unfortunately the traditional analyses of strength of cuts offer only limited help in understanding and addressing these issues''. We believe \gls{ML}/\gls{RL}  not only can be utilized  to achieve  improvements towards solving IPs in real applications, but may also aid researchers in understanding  effective selection of cutting planes. \added{While modern solvers use broader classes of cuts than just Gomory's, we decided to focus on Gomory's approach because it has the nice theoretical properties seen above, it requires no further input (e.g.~other human-designed cuts) and, as we will see, it leads to a well-defined and compact action space, and to clean evaluation criteria for the impact of \gls{RL}.}
 
 \paragraph{Our contributions.} We develop an \gls{RL} based method for intelligent adaptive selection of cutting planes, and use it in conjunction with Branch-and-Cut methods for efficiently solving \gls{IP}s. 
 Our main contributions are the following:
 \begin{itemize}[topsep=0pt,parsep=0pt,partopsep=0pt,leftmargin=*]
    \item {\bf Efficient MDP formulation.} We introduce an efficient \gls{MDP}  formulation for the problem of sequentially selecting cutting planes for an \gls{IP}. 
     Several trade-offs between the size of state-space/action-space vs.~generality of the method were navigated in order to arrive at the proposed formulation. For example, directly formulating the \gls{bnc} method as an \gls{MDP} would lead to a very large state space containing all open branches. Another example is the use of Gomory's cuts (vs.~other cutting plane methods), which helped limit the number of actions (available cuts)  in every round \added{to the number of variables. For some other classes of cuts, the number of available choices can be exponentially large.} 
     \item {\bf Deep \gls{RL} solution architecture design.} We build upon state-of-the-art deep learning techniques to design an efficient and scalable deep \gls{RL} architecture for learning to cut. Our design choices aim to address several unique challenges in this problem. These include slow state-transition machine (due to the complexity of solving \gls{LP}s) and the resulting need for an architecture that is easy to generalize\removed{parallelize}, order and size independent representation, reward shaping to handle frequent cases where the optimal solution is not reached, and handling numerical errors arising from the inherent nature of cutting plane methods. 
    \item {\bf Empirical evaluation.} 
   We evaluate our approach over  a range of classes of \gls{IP} problems (namely, packing, binary packing, planning, and maximum cut). Our experiments  demonstrate significant improvements in solution accuracy 
    as compared to popular human designed heuristics for adding Gomory's cuts.     Using our trained \gls{RL} policy for adding cuts in conjunction with \gls{bnc} methods leads to further significant improvements, thus illustrating the promise of our approach for improving state-of-the-art \gls{IP} solvers. 
     Further, we demonstrate the RL agent's potential to learn {\it meaningful} and effective cutting plane strategies through experiments on the well-studied \emph{knapsack} problem. In particular, we show that for this problem, the RL agent adds many more cuts resembling \emph{lifted cover inequalities} when compared to other heuristics. Those inequalities are well-studied in theory and known to work well for packing problems in practice. \removed{see~\cite{conforti2014integer,crowder1983solving,kellerer2003knapsack}.}
     Moreover, the RL agent is also shown to have  generalization properties across instance sizes and problem classes, in the sense that the RL agent trained on instances of one size or from one problem class is shown to perform competitively for instances of a different size and/or problem class.
     
 
 \end{itemize}

\section{Background on Integer Programming }\label{sec:background}

\paragraph{Integer programming.}It is well known that any Integer Program (\gls{IP}) can be written in the following \emph{canonical form}
\begin{equation}
                  \min \{ c^T x \, : \,  
                  Ax \leq b, \, 
                  x \geq 0, \,  x \in \mathbb{Z}^n\}
\label{eq:ip}
\end{equation}
where $x$ is the set of $n$ decision variables, $Ax \leq b, x \geq 0$ with $A\in \mathbb{Q}^{m\times n}, b\in \mathbb{Q}^m$ formulates the set of constraints, and the linear objective function is $c^T x$ for some $c \in \mathbb{Q}^n$. 
$x \in \mathbb{Z}^n$ implies we are only interested in integer solutions. Let $x^\ast_{\text{IP}}$ denote the optimal solution to the IP in~\eqref{eq:ip}, and $z^*_{\text{IP}}$ the corresponding objective value. 

\noindent {\bf The cutting plane method for integer programming.} The cutting plane method starts with solving the \gls{LP}  obtained from~\eqref{eq:ip} by\removed{on} dropping the integrality constraints $x \in \mathbb{Z}^n$. This LP is called the \emph{Linear Relaxation} (LR) of~\eqref{eq:ip}. Let ${\cal C}^{(0)} = \{x|Ax\leq b,x\geq 0\}$ be the feasible region of this LP, $x^\ast_{\text{LP}}(0)$ its optimal solution, and $z^*_{\text{LP}}(0)$ its objective value. Since ${\cal C}^{(0)}$ contains the feasible region of~\eqref{eq:ip}, we have  $z^*_{\text{LP}}(0)\leq z^*_{\text{IP}}$. Let us assume $x^\ast_{\text{LP}}(0) \notin \mathbb{Z}^n$. The cutting plane method then finds an inequality $a^T x\leq \beta$ (a \emph{cut}) that is satisfied by all integer feasible solutions of~\eqref{eq:ip}, but not by $x^\ast_{\text{LP}}(0)$ (one can prove that such an inequality always exists). The new constraint $a^T x \leq \beta$ is added to ${\cal C}^{(0)}$, to obtain  feasible region ${\cal C}^{(1)}\subseteq{\cal C}^{(0)}$; and then the new LP is solved, to obtain $x^\ast_{\text{LP}}(1)$. This procedure is iterated until $x^\ast_{\text{LP}}(t) \in \mathbb{Z}^n$. Since ${\cal C}^{(t)}$ contains the feasible region of~\eqref{eq:ip},  $x^\ast_{\text{LP}}(t)$ is an optimal solution to the integer program~\eqref{eq:ip}. In fact, $x^\ast_{\text{LP}}(t)$ is the  \emph{only} feasible solution to~\eqref{eq:ip} produced throughout the algorithm.

A typical way to compare cutting plane methods is by the number of cuts added throughout the algorithm: a better method is the one that terminates after adding a smaller number of cuts.  However, even for methods that are guaranteed to terminate in theory, in practice often numerical errors will prevent convergence to a feasible (optimal) solution. In this case, a typical way to evaluate the performance is the following. For an iteration $t$ of the method, the value $g^t:=z^*_{\text{IP}}-z^*_{\text{LP}}(t) \geq 0$ is called the (additive) \emph{integrality gap} of ${\cal C}^{(t)}$. Since ${\cal C}^{(t+1)}\subseteq {\cal C}^{(t)}$, we have that $g^t\geq g^{t+1}$. Hence, the integrality gap decreases during the execution of the cutting plane method. \added{A common way to} measure the performance of a cutting plane method is therefore given by computing the factor of integrality gap closed between the first LR, and the iteration $\tau$ when we decide to halt the method \added{(possibly without reaching an integer optimal solution)}, \added{see e.g.~\citet{wesselmann2012implementing}}. Specifically, we define the \gls{igc} as the ratio \begin{equation}\label{eq:igc}\frac{g^0-g^{\tau}}{g^0} \in [0,1].\end{equation} In order to measure the \gls{igc} achieved by \gls{RL}\removed{our learned} agent on test instances, we need to know the optimal value $z^*_{{\text{IP}}}$ for those instances, which we  compute with a commercial \gls{IP} solver. \added{Importantly, note that we do not use this measure, or the optimal objective value, for training, but only for evaluation. }

\noindent {\bf Gomory's Integer Cuts.} 
Cutting plane algorithms differ in how cutting planes are constructed at each iteration. Assume that the LR of~\eqref{eq:ip} with feasible region $\mathcal{C}^{(t)}$ has been solved via the simplex algorithm. 
At convergence, the simplex algorithm returns a so-called \emph{tableau}, which consists of a constraint matrix $\tilde{A}$ and a constraint vector $\tilde{b}$. Let $\mathcal{I}_t$ be the set of components  $[x^\ast_{\text{LP}}(t) ]_i$ that are fractional. For each $i\in \mathcal{I}_t$, we can generate a \emph{Gomory cut} \citep{gomory1960algorithm}
\begin{align}
(- \tilde{A}_{(i)} + \lfloor \tilde{A}_{(i)} \rfloor)^T x \leq - \tilde{b}_i + \lfloor \tilde{b}_i \rfloor,
\label{eq:cut}
\end{align}
where $\tilde{A}_{(i)}$ is the $i$th row of matrix $\tilde{A}$ and $\lfloor \cdot \rfloor$ means component-wise rounding down. Gomory cuts can therefore be generated for any \gls{IP} and, as required, are valid for all integer points from~\eqref{eq:ip} but not for $x^*_{\text{LP}}(t)$. 
Denote the set of all candidate cuts in round $t$ as $\mathcal{D}^{(t)}$, so that $I_t \coloneqq  |\mathcal{D}^{(t)}| = |\mathcal{I}_t| $.

It is shown in \citet{gomory1960algorithm} that a cutting plane method which at each step adds \added{an appropriate} Gomory's cut terminates in a finite number of iteration. At each iteration $t$, we have as many as $I_t \in [n]$ cuts to choose from. As a result, the efficiency and quality of the solutions depend highly on the sequence of generated cutting planes, which are usually chosen by heuristics~\citep{wesselmann2012implementing}. We aim to show that the \added{choice of Gomory's cuts, hence the quality of the solution,} can be significantly improved with \gls{RL}.\removed{an adaptive \gls{RL} based cut selection. }

\noindent {\bf Branch and cut.} In state-of-the-art solvers, the addition of cutting planes is alternated with a \emph{branching phase}, which can be described as follows. Let $x^\ast_{\text{LP}}(t)$ be the solution to the current LR of~\eqref{eq:ip}, and assume that some component of $x^\ast_{\text{LP}}(t)$, say wlog the first, is not integer (else, $x^\ast_{\text{LP}}(t)$ is the optimal solution to~\eqref{eq:ip}). Then~\eqref{eq:ip} can be split into two \emph{subproblems}, whose LRs are obtained from ${\cal C}^{(t)}$ by adding constraints $x_1 \leq \lfloor [x^\ast_{\text{LP}}(t)]_1\rfloor$ and $x_1 \geq \lceil [x^\ast_{\text{LP}}(t)]_1\rceil$, respectively. Note that the set of feasible integer points for~\eqref{eq:ip} is the union of the set of feasible integer points for the two new subproblems. Hence, the integer solution with minimum value (for a minimization IP) among those  subproblems gives the optimal solution to~\eqref{eq:ip}. Several heuristics are used to select which subproblem to solve next, in attempt to minimize the number of suproblems (also called \emph{child nodes}) created. An algorithm that alternates between the cutting plane method and branching is called \emph{Branch-and-Cut} (\gls{bnc}). When all the other parameters (e.g., the number of cuts added to a subproblem) are kept constant, a typical way to evaluate a B\&C method is by the number of subproblems explored before the optimal solution is found.  


\section{Deep {RL} Formulation and Solution Architecture}
Here we present our formulation of the  cutting plane selection problem as an \gls{RL} problem, and our deep RL based solution architecture. \removed{our deep \gls{RL} network architecture for solving the same.}

\subsection{Formulating Cutting Plane selection as {RL}}
The standard \gls{RL} formulation starts with an \gls{MDP}: at time step $t \geq 0$, an agent is in a state $s_t \in \mathcal{S}$, takes an action $a_t \in \mathcal{A}$, receives an instant reward $r_t \in \mathbb{R}$ and transitions to the next state $s_{t+1} \sim p(\cdot|s_t,a_t)$. A policy  $\pi:\mathcal{S} \mapsto \mathcal{P}(\mathcal{A})$ gives a mapping from any state to a distribution over actions $\pi(\cdot|s_t)$. The objective of \gls{RL} is to search for a policy that maximizes the expected cumulative rewards over a horizon $T$, i.e., 
$
    \max_\pi \ J(\pi) := \mathbb{E}_\pi[\sum_{t=0}^{T-1} \gamma^tr_t]
$,
where $\gamma \in (0,1]$ is a discount factor and the expectation is w.r.t. randomness in the policy $\pi$ as well as the environment (e.g.~the transition dynamics $p(\cdot|s_t,a_t)$). In practice, we consider  parameterized policies $\pi_\theta$ 
and aim to find $\theta^\ast = \arg\max_\theta J(\pi_\theta)$. Next, we formulate the procedure of selecting cutting planes into an \gls{MDP}. We specify state space $\mathcal{S}$, action space $\mathcal{A}$, reward $r_t$ and the transition $s_{t+1}\sim p(\cdot|s_t,a_t)$.

\noindent{\bf State Space $\mathcal{S}$.} At iteration $t$, the new \gls{LP} is defined by the feasible region  $\mathcal{C}^{(t)}=\{a_i^T x \leq b_i\}_{i=1}^{N_t}$ where $N_t$ is the total number of constraints  including the original linear constraints (other than non-negativity) in the IP and the cuts added so far. Solving the resulting \gls{LP} produces an optimal solution $x_{\text{LP}}^\ast(t)$ along with the set of candidate Gomory's cuts $\mathcal{D}^{(t)}$. We set the numerical representation of the state to be $s_t = \{\mathcal{C}^{(t)},c,x_{\text{LP}}^\ast(t),\mathcal{D}^{(t)}\}$.  When all components of  $x_{\text{LP}}^\ast(t)$ are integer-valued, $s_t$ is a terminal state and $\mathcal{D}^{(t)}$ is an empty set.

\noindent{\bf Action Space $\mathcal{A}$.} At iteration $t$, the available actions are given by $\mathcal{D}^{(t)}$, consisting of all possible  Gomory's cutting planes that can be added to the \gls{LP} in the next iteration. 
The action space is discrete because each action is a discrete choice of the cutting plane. 
 However, each action is represented as an inequality\removed{(the cutting plane)} $e_i^T x\leq d_i$, and therefore is parameterized by  $e_i \in \mathbb{R}^n, d_i\in \mathbb{R}$. This is different from conventional discrete action space which can be an arbitrary unrelated set of actions.

\noindent{\bf Reward $r_t$.}\removed{We encourage the cutting planesto cut aggressively.} \added{To encourage adding cutting plane aggressively,} \removed{We therefore} \added{we} set the instant reward in iteration $t$ to be the gap between objective values of consecutive LP solutions, that is, $r_t = c^T x_{\text{LP}}^\ast(t+1) -  c^T x_{\text{LP}}^\ast(t) \geq 0$. With a discount factor $\gamma < 1$, this encourages the agent to reduce the integrality gap and approach the integer optimal solution as fast as possible. 


\noindent{\bf Transition.} 
Given state $s_t = \{\mathcal{C}^{(t)},c,x_{\text{LP}}^\ast(t),\mathcal{D}^{(t)}\}$, on taking action $a_t$ (i.e., on adding a  chosen cutting plane $e_i^Tx \leq d_i$), the new state $s_{t+1}$ is determined as follows. Consider the new constraint set $\mathcal{C}^{(t+1)} = \mathcal{C}^{(t)} \cup \{e_i^T x\leq d_i\}$. The augmented set of constraints $\mathcal{C}^{(t+1)}$ form a new \gls{LP}, which can be efficiently solved using the simplex method to get $x_{\text{LP}}^\ast(t+1)$. The new set of Gomory's cutting planes $\mathcal{D}^{(t+1)}$ can then be computed from the simplex tableau. 
Then, the new state $s_{t+1} = \{\mathcal{C}^{(t+1)},c,x_{\text{LP}}^\ast(t+1),\mathcal{D}^{(t+1)}\}$.


 
\subsection{Policy Network Architecture}
We now describe the policy network architecture for $\pi_\theta(a_t|s_t)$. Recall from the last section we have in the state $s_t$ a set of inequalities $\mathcal{C}^{(t)} = \{a_i^Tx \leq b_i\}_{i=1}^{N_t}$, and as available actions, another set  $\mathcal{D}^{(t)} =\{e_i^Tx \leq d_i\}_{i=1}^{I_t}$. Given state $s_t$, a policy $\pi_\theta$ specifies a distribution over $\mathcal{D}^{(t)}$, via the following architecture.

\noindent{\bf Attention network for order-agnostic cut selection.} Given current LP constraints in $\mathcal{C}^{(t)}$, when computing distributions over the $I_t$ candidate constraints in $\mathcal{D}^{(t)}$, it is desirable that the architecture is agnostic to the ordering among the constraints (both in $\mathcal{C}^{(t)}$ and $\mathcal{D}^{(t)}$), 
because the ordering does not reflect the geometry of the feasible set. To achieve this, we adopt ideas from \added{the}\removed{the work on} attention network \citep{vaswani2017attention}. \removed{Under this approach, we}\added{We} use a parametric function $F_\theta:\mathbb{R}^{n+1} \mapsto \mathbb{R}^k$ for some given $k$ (encoded by a \removed{DNN}\added{network} with parameter $\theta$). This function is used to compute projections $h_i = F_\theta([a_i,b_i]), i \in [N_t]$ and $g_j = F_\theta([e_j,d_j]),j\in [I_t]$ for each inequality in $\mathcal{C}^{(t)}$ and $\mathcal{D}^{(t)}$, respectively. Here $[\cdot,\cdot]$ \added{denotes}\removed{represents} concatenation. The score $S_j$ for every candidate cut $j\in [I_t]$ is \removed{then} computed as  
\begin{equation}
    \textstyle
    S_j = \frac{1}{N_t} \sum_{i=1}^{N_t} g_j^T h_i
    \label{eq:attention}
\end{equation}
Intuitively, when assigning these scores  to the candidate cuts, (\ref{eq:attention})  accounts for each candidate's interaction with all the constraints already in the \removed{system}\added{\gls{LP}} through the inner products $g_j^T h_i$. 
We then define probabilities $p_1, \ldots, p_{I_t}$ by a softmax function $\text{softmax}(S_1, \ldots, S_{I_t})$. The resulting $I_t$-way categorical distribution is the distribution over actions given by policy $\pi_\theta(\cdot|s_t)$ in the current state $s_t$. 

\noindent {\bf LSTM network for variable sized inputs.} 
We want our RL agent to be able to handle IP instances of different sizes (number of decision variables and constraints). 
Note that the number of constraints can vary over different iterations of a cutting plane method even for a fixed \gls{IP} instance. But this variation is not a concern since the attention network \removed{architecture} described above \removed{is able to handle}\added{handles} that variability in a natural way. To be able to use the same policy network for instances with different number of variables \removed{using}, we embed each constraint using a LSTM network $\text{LSTM}_\theta$ \citep{hochreiter1997long} with hidden state of size $n+1$ for a fixed $n$. In particular, for a general constraint $\tilde{a}_i^T \tilde{x} \leq \tilde{b}_i$ with $\tilde{a}_i\in \mathbb{R}^{\tilde{n}}$ with $\tilde{n} \neq n$, we carry out the embedding
$\tilde{h}_i = \text{LSTM}_\theta([\tilde{a}_i,\tilde{b}_i])$
where $\tilde{h}_i \in \mathbb{R}^{n+1}$ is the last hidden state of the LSTM network. This hidden state $\tilde{h}_i$ can be used in place of $[\tilde{a}_i,\tilde{b}_i]$ in the attention network. The idea is that the hidden state $\tilde{h}_i$ can properly encode all information in the original inequalities $[\tilde{a}_i,\tilde{b}_i]$ if the LSTM network is powerful enough.

\added{{\noindent \bf Policy rollout.} To put everything together, in Algorithm~\ref{algo:routine}, we lay out the steps involved in rolling out a policy, i.e., executing a  policy on a given IP instance.} 

 \begin{algorithm}[t]
	\begin{algorithmic}[1]
		\STATE Input: policy network parameter $\theta$, \gls{IP} instance parameterized by $c, A, b$, number of iterations $T$.
		\STATE Initialize iteration counter $t = 0$.
		\STATE Initialize minimization LP with constraints $\mathcal{C}^{(0)} = \{Ax \leq b\}$ and cost vector $c$. Solve to obtain $x_{\text{LP}}^\ast(0)$. Generate set of candidate cuts $\mathcal{D}^{(0)}$.
		\WHILE { $x_{\text{LP}}^\ast(t)$ not all integer-valued and $t \leq T$}
		\STATE Construct state $s_t = \{\mathcal{C}^{(t)},c,x_{\text{LP}}^\ast(t),\mathcal{D}^{(t)}\}$. 
		\STATE Sample an action using the  distribution over candidate cuts given by policy $\pi_\theta$, as $a_t \sim \pi_\theta(\cdot|s_t)$. Here the action $a_t$ corresponds to a cut $\{e^T x\leq d\} \in \mathcal{D}^{(t)}$.
		\STATE  Append the cut to the constraint set, $\mathcal{C}^{(t+1)} = \mathcal{C}^{(t)} \cup\{e^T x\leq d\} $. Solve for $x_{\text{LP}}^\ast(t+1)$. 
		Generate $\mathcal{D}^{(t+1)}$.
		\STATE Compute reward $r_t$.
		\STATE $t \leftarrow t + 1$.
		\ENDWHILE
	\end{algorithmic}
	 	\caption{Rollout of the Policy}\label{algo:routine}
\end{algorithm}

\subsection{Training: Evolutionary Strategies} 
We train the \gls{RL} agent using \gls{ES} \citep{salimans2017evolution}. The core idea \removed{behind this technique} is to flatten the \gls{RL} problem into a blackbox optimization problem where the input is a policy parameter $\theta$ and the output is a noisy estimate of the agent's performance under the corresponding policy. \gls{ES} apply random sensing to approximate the policy gradient $\hat{g}_\theta \approx \nabla_\theta J(\pi_\theta)$ and then carry out the iteratively update $\theta \leftarrow \theta + \hat{g}_\theta$ for some $\alpha >  0$.  The gradient estimator takes the following form 
\begin{equation}
\textstyle
    \hat{g}_\theta = \frac{1}{N} \sum_{i=1}^N J(\pi_{\theta_i^\prime}) \frac{\epsilon_i}{\sigma}, 
    \label{eq:esgrad}
\end{equation}
where $\epsilon_i \sim \mathcal{N}(0,\mathbb{I})$ is a sample from a multivariate Gaussian, $\theta_i^\prime = \theta + \sigma\epsilon_i$ and $\sigma>0$ is a fixed constant. Here the return $J({\pi_{\theta^\prime}})$ can be estimated as $ \sum_{t=0}^{T-1} r_t\gamma^t$ using a single trajectory (or average over multiple trajectories) generated on executing the policy  $\pi_{\theta^\prime}$, as in Algorithm \ref{algo:routine}. 
To train the policy on $M$ distinct \gls{IP} instances, we average the \gls{ES} gradient estimators over all instances. \removed{\paragraph{Why ES.}} \added{Optimizing the policy with \gls{ES} comes with several advantages, e.g., simplicity of communication protocol between workers when compared to some other actor-learner based distributed algorithms \citep{espeholt2018impala,kapturowski2018recurrent}, and simple parameter updates.} \removed{especially when state/action space are irregular: \textbf{(1)} Communication: Compared to some other actor-learner based distributed algorithms \citep{espeholt2018impala,kapturowski2018recurrent}, \gls{ES} train with very simple  protocols between workers; \textbf{(2)} Simple Updates: Parameter updates for \gls{ES} are much simpler because no batched back-propagation is needed.}Further \removed{details and} discussions are in the appendix.

\subsection{Testing}
\added{We test the performance of a trained policy $\pi_\theta$ by rolling out (as in Algorithm \ref{algo:routine}) on a set of test instances, and measuring the \gls{igc}.} 
\removed{We can test the performance of a trained policy $\pi_\theta$ for a given \gls{IP} instance, by executing it as described in Algorithm \ref{algo:routine}.}
One important design consideration is that a cutting plane method can potentially cut off the optimal integer solution due to the \gls{LP} solver's numerical errors. Invalid cutting planes generated by numerical errors is a well-known phenomenon in integer programming~\cite{CornuejolsMargot}. Further, learning can amplify this problem. This is because an RL policy trained to \added{decrease the cost of the LP solution} 
might learn to aggressively add cuts in order to tighten the \gls{LP} constraints. When no countermeasures were taken, 
we observed that the \gls{RL} agent could cut the optimal solution in as many as 20$\%$ of the instances for some problems! 
To remedy this, we have added a simple \emph{stopping criterion} at test time. The idea is to maintain a running statistics that measures the relative progress made by newly added cuts during execution. When a certain number of consecutive cuts have little effect on the \gls{LP} objective value, we simply terminate the episode. This prevents the agent from adding cuts that are likely to induce numerical errors. Indeed, our \removed{computational} experiments show this modification is enough to completely remove the generation of invalid cutting planes. We postpone the details to the appendix. 




\section{Experiments}
We evaluate our approach \added{with}\removed{using} a variety of experiments, designed to examine the {\it quality} of the cutting plane\added{s} \removed{algorithm}\removed{learned}\added{selected} by \gls{RL}\removed{(in the form of a policy network)}. Specifically, we conduct five sets of experiments to evaluate our approach from the \removed{following} different aspects:
\begin{enumerate}[leftmargin=*]
    \item {\bf Efficiency of cuts.} Can the \gls{RL} agent solve an \gls{IP} problem using fewer number of Gomory cuts?
    \item {\bf Integrality gap closed.} 
    \removed{For larger instances, where}\added{In cases where} cutting plane\added{s} \removed{methods} alone are unlikely to solve the problem to optimality, can the \gls{RL} agent close the integrality gap effectively?
    \removed{As discussed in Section \ref{sec:background}, for larger \gls{IP} problems cutting plane methods alone are unlikley to solve the problem to optimality. For such problems, can the \gls{RL} agent be used to close the integrality gap effectively? We use the metric of  \gls{igc} defined in \eqref{eq:igc}, which measures the factor of integrality gap closed compared to the initial gap. }
    \item {\bf Generalization properties.}
        \begin{itemize}
        \item (size) Can an \gls{RL} agent trained on smaller instances be applied to 10X larger instances to yield performance 
        comparable to an agent trained on the larger instances?  
        \item (structure) Can an \gls{RL} agent trained on instances from one class of \gls{IP}s be applied to a very different class of \gls{IP}s to  yield performance 
        comparable to an agent trained on the latter class?  
        \end{itemize}
    \item {\bf Impact on the efficiency of \gls{bnc}.} \removed{As mentioned in Section~\ref{sec:background}, in practice, cutting planes are alternated with a branching procedure, leading to Branch-and-Cut (\gls{bnc}).  }
   \added{Will the \gls{RL} agent trained as a cutting plane method be effective as a subroutine within a \gls{bnc} method?}  
   \removed{We evaluate the impact on the efficiency of \gls{bnc} in terms of reduction in the number of subproblems (aka nodes) created until an optimal solution is reached \added{(or, for larger instances, until IGC is close to $1$)}.}
   \item {\bf Interpretability of cuts: the knapsack problem.}
   Does RL have \added{the} potential to provide insights into effective and meaningful cutting plane strategies for specific problem\added{s}? \added{Specifically, for the knapsack problem, do the cuts learned by RL resemble lifted cover inequalities?} \removed{To gain some insight into this aspect, we perform an interesting set of experiments for the knapsack problem. For this very fundamental class of \gls{IP} problems, we focus on cuts of a specific form (lifted cover inequalities), that are known to work well in theory and practice. We study if the cuts learned by RL resemble these well-known inequalities.}

\end{enumerate}



\noindent{\bf IP instances used for training and testing.} We consider four classes of \gls{IP}s: Packing, Production Planning, Binary Packing and Max-Cut. These represent a wide collection of well-studied \gls{IP}s ranging from resource allocation to graph optimization. The \gls{IP} formulations of these problems are provided in the appendix. \removed{We let}\added{Let} $n,m$ denote the number of variables and constraints (other than nonnegativity) in the \gls{IP} formulation, so that $n\times m$ denotes the size of the \gls{IP} instances \added{(see tables below)}\removed{in the result tables for different experiments}. The mapping from specific problem parameters (like number of nodes and edges in maximum-cut) to $n,m$ depends on the \gls{IP} formulation used for each problem. 
\added{We use randomly generated problem instances for training and testing the RL agent for each \gls{IP} problem class. For the small ($n\times m \approx 200$) and medium ($n\times m \approx 1000$) sized problems we used $30$ training instances and $20$ test instances. These numbers were doubled for larger problems ($n\times m \approx 5000$).}
\added{Importantly, note that we do not need ``solved" (aka labeled) instances for training. \gls{RL} only requires repeated rollouts on training instances. }

\noindent{\bf Baselines.} We compare the performance of the RL agent with the following commonly used human-designed heuristics for choosing (Gomory) cuts \citep{wesselmann2012implementing}: Random, Max Violation (MV), Max Normalized Violation (MNV) and Lexicographical Rule (LE)\added{, with LE being the original rule used in Gomory's method, for which a theoretical convergence in finite time is guaranteed}. Precise descriptions of these heuristics are in the appendix. 

\noindent{\bf Implementation details.} We implement the \gls{MDP} simulation environment for \gls{RL} using  Gurobi \citep{gurobi2015gurobi} as the \gls{LP} solver. The C\removed{-programming} interface of Gurobi \removed{allows us to efficiently add a new constraint}\added{entails efficient addition of new constraints} (i.e., the cut chosen by \gls{RL} agent) to the current LP and solve the modified \gls{LP}. 
The number of cuts added (i.e., the horizon $T$ in rollout of a policy)  depend on the problem size.  We sample actions from the categorical distribution $\{p_i\}$ during training; but during testing, we take actions greedily as $i^\ast = \arg\max_i p_i$. Further  implementation details, along with hyper-parameter settings \added{for the \gls{RL} method are provided in the appendix.}\removed{for the policy network and the \gls{ES} method, are provided in the appendix.}
\setcounter{figure}{1}

\paragraph{Experiment \#1: Efficiency of cuts (small-sized instances).}
For small-sized \gls{IP} instances, cutting planes alone can potentially solve an IP problem to optimality. For such instances, we compare different cutting plane methods on the total number of cuts it takes to find an optimal integer solution. 
\added{ Table \ref{table:smallsize} shows that the RL agent achieves close to several factors of improvement in the number of cuts required, 
when compared to the baselines. Here,  for each class of IP problems, the second row of the table gives the size of the IP formulation of the instances used for training and testing.}

\begin{table}[t]
\caption{\small{Number of cuts it takes to reach optimality. We show $\text{mean} \pm \text{std}$ across all test instances.}}
\begin{small}
\begin{sc}
\begin{tabular}{C{0.37in} *4{C{.5in}}}\toprule[1.5pt]
\bf Tasks & \bf Packing  & \bf Planning & \bf Binary & \bf Max Cut \\\midrule
\bf Size & $10 \times 5$  & $13 \times 20$ & $10 \times 20$ &  $10 \times 22$  \\\midrule 
Random      &  $48  \pm 36$  & $44 \pm 37$ & $81 \pm 32$ & $69 \pm 34$  \\
MV & $62 \pm 40$ & $48 \pm 29$ & $87 \pm 27$ & $64 \pm 36$  \\ 
MNV & $53 \pm 39$ & $60 \pm 34$ & $85 \pm 29$ &$47 \pm 34$ \\ 
LE & $34 \pm 17$ & $310 \pm 60$ & $89 \pm 26$ &$59 \pm 35$ \\ 
RL & $\mathbf{14 \pm 11}$ & $\mathbf{10 \pm 12}$ & $\mathbf{22 \pm 27}$ & $\mathbf{13 \pm 4}$ \\
\bottomrule
\end{tabular}
\end{sc}
\end{small}
\vskip -0.1in
\label{table:smallsize}
\end{table}

\begin{table}[t!]
\caption{\small{\gls{igc} for test instances of size roughly $1000$. We show $\text{mean} \pm \text{std}$ of \gls{igc} achieved on adding $T=50$ cuts.}}
\begin{small}
\begin{sc}
\begin{tabular}{C{0.18in} *4{C{.56in}}}\toprule[1.5pt]
\bf Tasks & \bf Packing  & \bf Planning & \bf Binary & \bf Max Cut \\\midrule
\bf Size & $30 \times 30$  & $61 \times 84$ & $33 \times 66$ & $27 \times 67$  \\\midrule
Rand      &  $0.18 \pm 0.17$  & $0.56 \pm 0.16$ & $0.39 \pm 0.21$ & $0.56 \pm 0.09$ \\
MV & $0.14 \pm 0.08$ & $0.18 \pm 0.08$ & $0.32 \pm 0.18$ & $0.55 \pm 0.10$  \\ 
MNV & $0.19 \pm 0.23$ & $0.31 \pm 0.09$ & $0.32 \pm 0.24$ & $0.62 \pm 0.12$ \\ 
LE & $0.20 \pm 0.22$ & $0.01 \pm 0.01$ & $0.41 \pm 0.27$ & $0.54 \pm 0.15$ \\
RL & $\mathbf{0.55 \pm 0.32}$ & $\mathbf{0.88 \pm 0.12}$ & $\mathbf{0.95 \pm 0.14}$ & $\mathbf{0.86 \pm 0.14}$ \\
\bottomrule
\end{tabular}
\end{sc}
\end{small}
\label{table:optimalitygap_medium}
\end{table}

\begin{figure}[t!]
\centering
\subfigure[Packing]{\includegraphics[width=.48\linewidth]{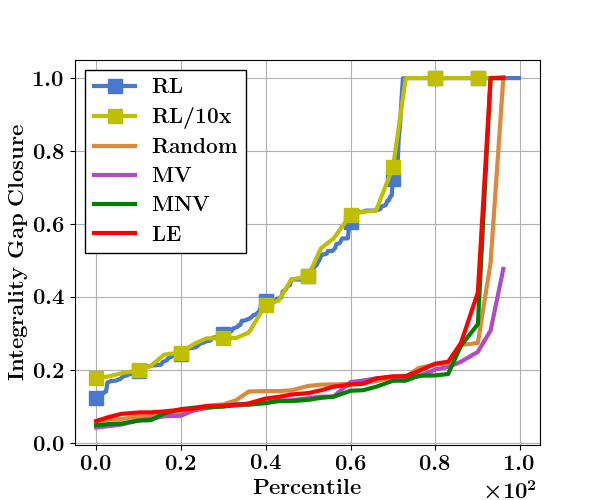}}
\subfigure[Planning]{\includegraphics[width=.48\linewidth]{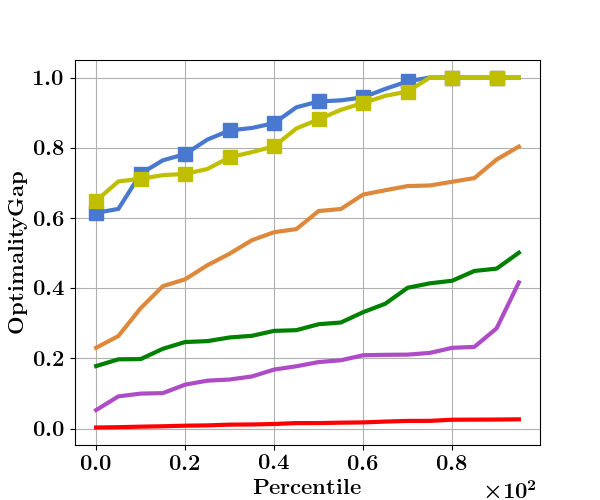}}
\subfigure[Binary Packing]{\includegraphics[width=.48\linewidth]{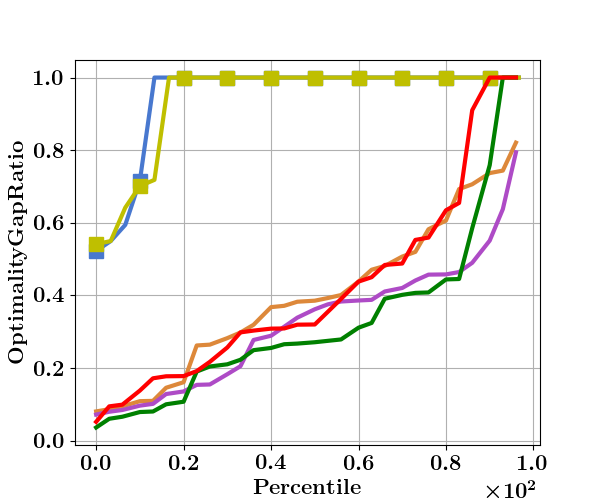}}
\subfigure[Max Cut]{\includegraphics[width=.48\linewidth]{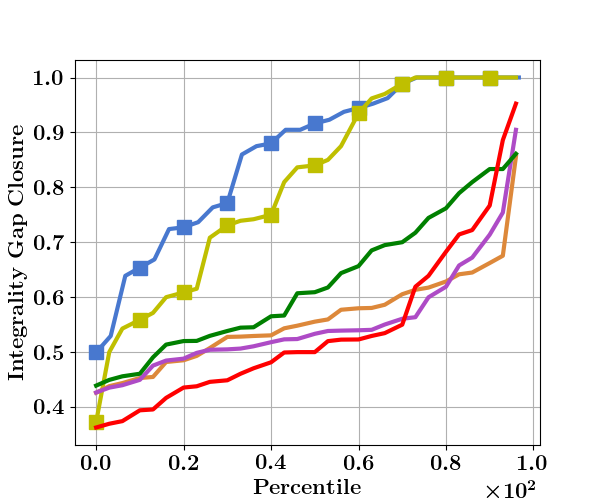}}
\caption{\small{Percentile plots of IGC for test instances of size roughly $1000$. X-axis shows the percentile of instances and y-axis shows the \gls{igc} achieved on adding $T=50$ cuts. Across all test instances, \gls{RL} achieves significantly higher IGC than the baselines.}}
\label{figure:objective_medium}
\end{figure}

\begin{figure}[t!]
\centering
\subfigure[Packing]{\includegraphics[width=.48\linewidth]{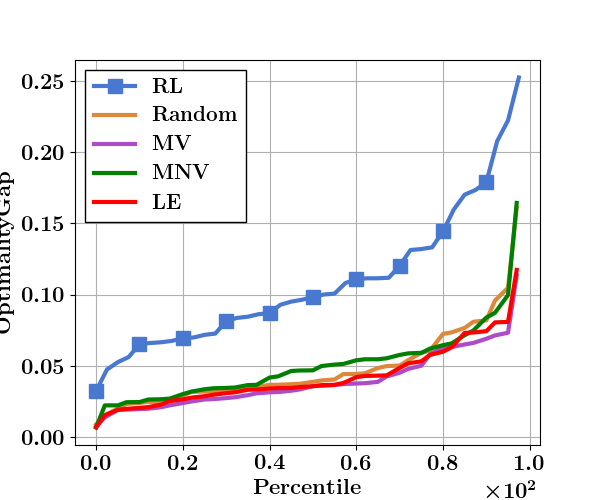}}
\subfigure[Planning]{\includegraphics[width=.48\linewidth]{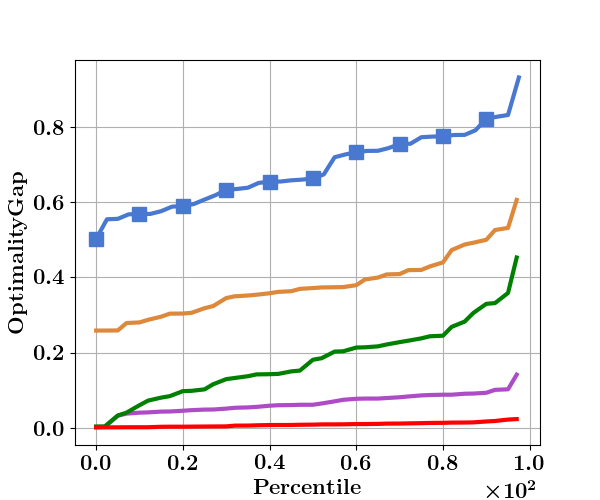}}
\subfigure[Binary Packing]{\includegraphics[width=.48\linewidth]{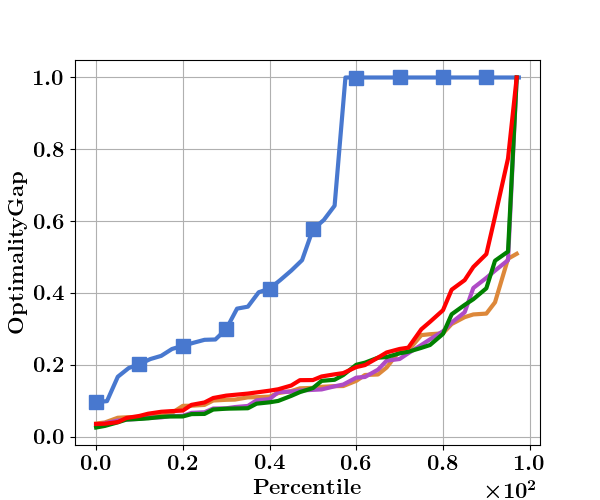}}
\subfigure[Max Cut]{\includegraphics[width=.48\linewidth]{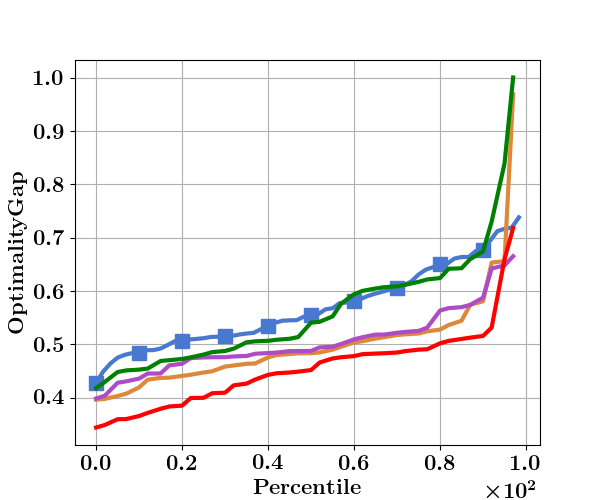}}
\caption{\small{Percentile plots of IGC for test instances of size roughly $5000$, $T=250$ cuts. Same set up as Figure \ref{figure:objective_medium}  but on even larger-size instances.  }}
\label{figure:objective_largescale}
\end{figure}

\paragraph{Experiment \#2: Integrality gap closure for large-sized instances.}
Next, we train and test the RL agent on significantly larger problem instances compared to the previous experiment. 
\added{In the first set of experiments (Table \ref{table:optimalitygap_medium} and Figure \ref{figure:objective_medium}), we consider instances of size ($n\times m$) close to $1000$. In Table \ref{table:optimalitygap_largescale} and Figure \ref{figure:objective_largescale}, we report results for even larger scale problems, with instances of size close to $5000$.
We add $T=50$ cuts for the first set of instances, and $T=250$ cuts for the second set of instances. 
However, for these instances, the cutting plane methods is unable to reach optimality. Therefore, we compare different cutting plane methods on integrality gap closed using the IGC metric defined in \eqref{eq:igc}, Section \ref{sec:background}. 
Table \ref{table:optimalitygap_medium},  \ref{table:optimalitygap_largescale} show that on average RL agent was able to close a significantly higher fraction of gap compared to the other methods.  Figure \ref{figure:objective_medium}, \ref{figure:objective_largescale} provide a more detailed comparison, by showing a percentile plot -- here the instances are sorted in the ascending order of \gls{igc}  and then plotted in order; the $y$-axis shows the \gls{igc} and the $x$-axis shows the percentile of instances achieving that \gls{igc}. The blue curve with square markers shows the performance of our RL agent. In Figure \ref{figure:objective_medium}, very close to the blue curve is the yellow curve (also with square marker). This yellow curve is for RL$/$10X, which is an RL agent trained on 10X smaller instances in order to evaluate generalization properties, as we describe next. }

\begin{table}[t!]
\caption{\small{IGC for test instances of size roughly $5000$. We show $\text{mean} \pm \text{std}$ of \gls{igc} achieved on adding $T=250$ cuts. }}
\begin{small}
\begin{sc}
\begin{tabular}{C{0.35in} *4{C{.56in}}}\toprule[1.5pt]
\bf Tasks & \bf Packing  & \bf Planning & \bf Binary & \bf Max Cut \\\midrule
\bf Size & $60 \times 60$  & $121 \times 168$ & $66 \times 132$ & $54 \times 134$  \\\midrule
Random      &  $0.05 \pm 0.03$  & $0.38\pm 0.08$ & $0.17 \pm 0.12$ & $0.50 \pm 0.10$ \\
MV & $0.04 \pm 0.02$ & $0.07 \pm 0.03$ & $0.19 \pm 0.18$ & $0.50 \pm 0.06$  \\ 
MNV & $0.05 \pm 0.03$ & $0.17 \pm 0.10$ & $0.19 \pm 0.18$ & $0.56 \pm 0.11$ \\ 
LE & $0.04 \pm 0.02$ & $0.01 \pm 0.01$ & $0.23 \pm 0.20$ & $0.45 \pm 0.08$ \\
RL & $\mathbf{0.11 \pm 0.05}$ & $\mathbf{0.68 \pm 0.10}$ & $\mathbf{0.61 \pm 0.35}$ & $\mathbf{0.57 \pm 0.10}$ \\
\bottomrule
\end{tabular}
\end{sc}
\end{small}
\label{table:optimalitygap_largescale}
\end{table}

\begin{figure}[t]
\centering
\includegraphics[width=0.4\textwidth]{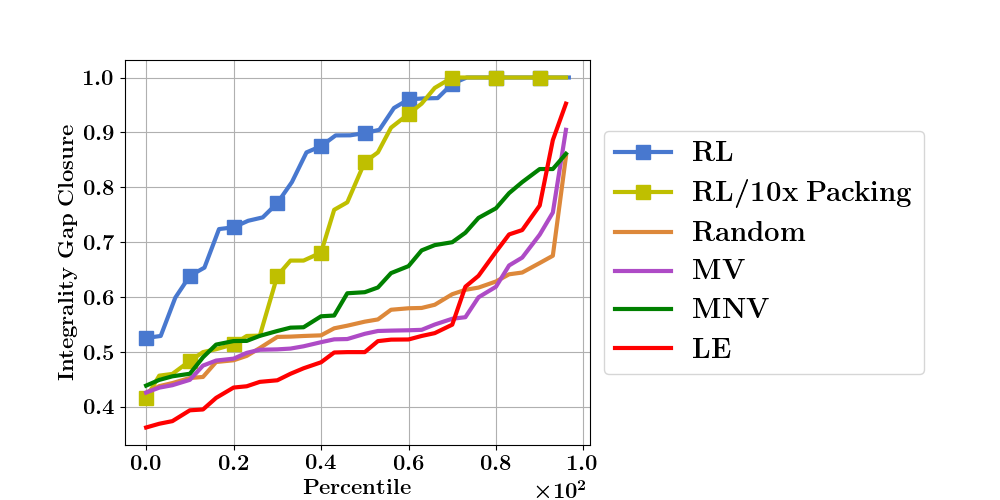}
\caption{{\small
Percentile plots of Integrality Gap Closure. `RL/10X packing' trained on instances of a completely different IP problem (packing)  performs competitively on the maximum-cut instances.}}
 \label{figure:crossgeneralization}
\end{figure}

\begin{table}[t]
\caption{\small{IGC in \gls{bnc}. We show $\text{mean} \pm \text{std}$ across test instances.}}
\begin{small}
\begin{sc}
\begin{tabular}{C{0.4in} *4{C{.56in}}}\toprule[1.5pt]
\bf Tasks & \bf Packing  & \bf Planning & \bf Binary & \bf Max Cut \\\midrule
\bf Size & \bf $30 \times 30$  & \bf $61 \times 84$ & \bf $33 \times 66$ & \bf $27 \times 67$  \\\midrule
No cut   &  $0.57 \pm 0.34$  & $0.35 \pm 0.08$ & $0.60 \pm 0.24$ & $1.0 \pm 0.0$ \\
Random      &  $0.79 \pm 0.25$  & $0.88 \pm 0.16$ & $0.97 \pm 0.09$ & $1.0 \pm 0.0$ \\
MV & $0.67 \pm 0.38$ & $0.64 \pm 0.27$ & $0.97 \pm 0.09$ & $0.97 \pm 0.18$  \\ 
MNV & $0.83 \pm 0.23$ & $0.74 \pm 0.22$ & $1.0 \pm 0.0$ & $1.0 \pm 0.0$ \\ 
LE & $0.80 \pm 0.26$ & $0.35 \pm 0.08$ & $0.97 \pm 0.08$& $1.0 \pm 0.0$ \\ 
RL & $\mathbf{0.88 \pm 0.23}$ & $\mathbf{1.0 \pm 0.0}$ & $1.0 \pm 0.0$ & $1.0 \pm 0.0$ \\
\bottomrule
\end{tabular}
\end{sc}
\end{small}
\vskip -0.1in
\label{table:numberofnodes}
\end{table}
\begin{figure}[h!]
\centering
\subfigure[Packing (1000 nodes)]{\includegraphics[width=.48\linewidth]{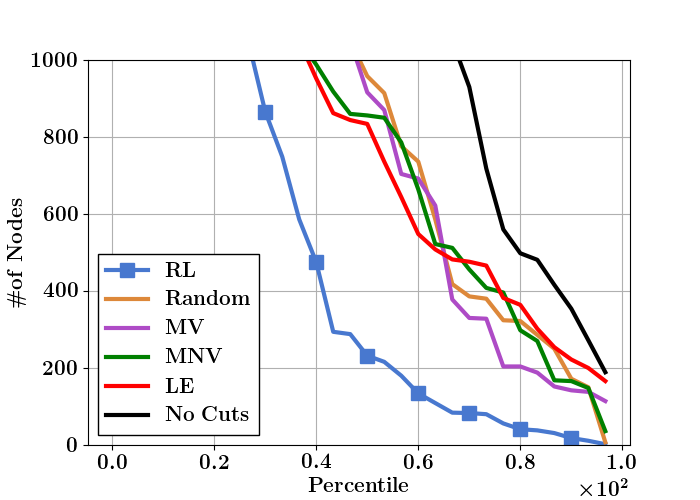}}
\subfigure[Planning (200 node)]{\includegraphics[width=.48\linewidth]{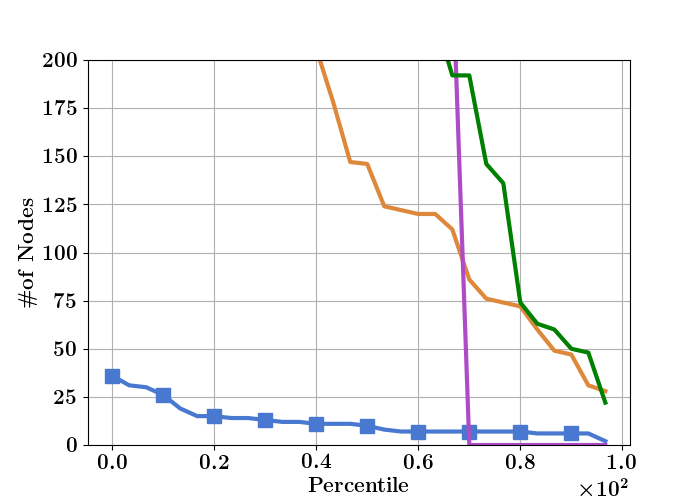}}
\subfigure[Binary (200 nodes) ]{\includegraphics[width=.48\linewidth]{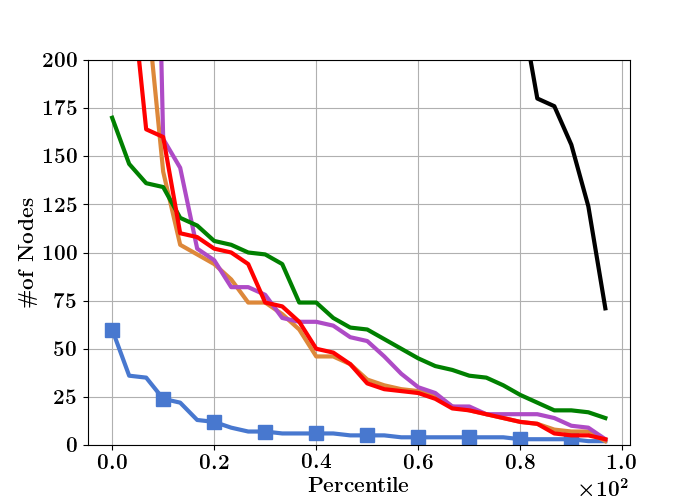}}
\subfigure[Max Cut (200 nodes)]{\includegraphics[width=.48\linewidth]{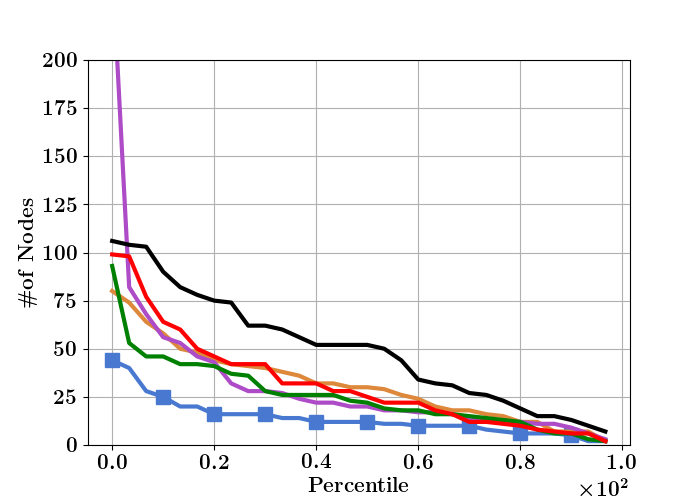}}
\caption{\small{Percentile plots of number of B\&C nodes expanded. X-axis shows the percentile of instances and y-axis shows the number of expanded nodes to close $95\%$ of the integrality gap.}}
\label{figure:numberofbb}
\end{figure}

\begin{figure*}[t!]
\centering
\subfigure[Criterion 1]{\includegraphics[width=.24\linewidth]{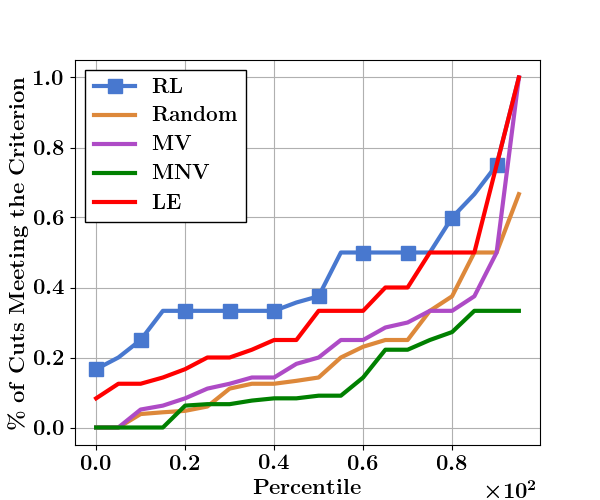}}
\subfigure[Criterion 2]{\includegraphics[width=.24\linewidth]{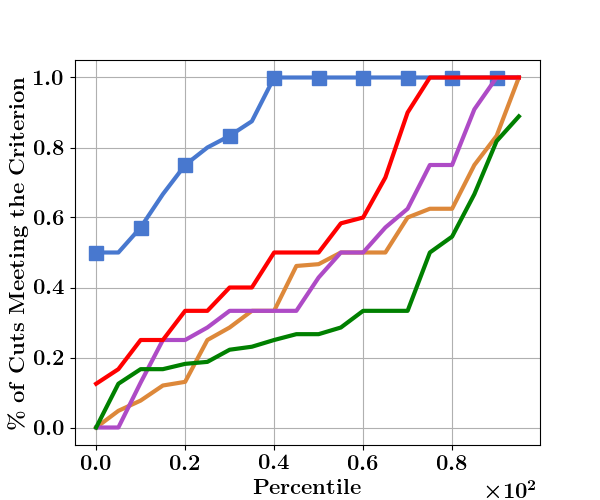}}
\subfigure[Criterion 3]{\includegraphics[width=.24\linewidth]{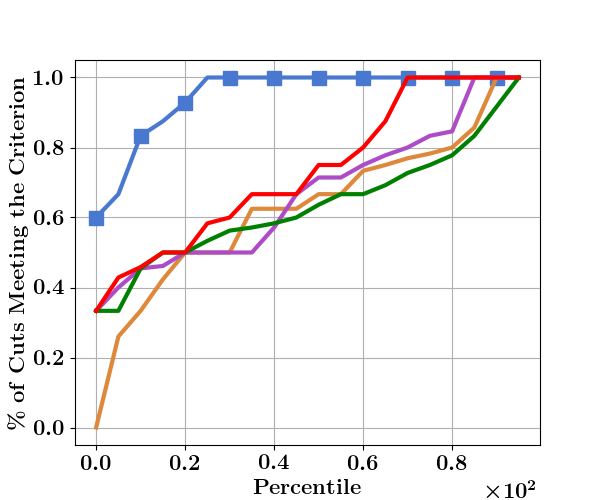}}
\subfigure[Number of cuts]{\includegraphics[width=.24\linewidth]{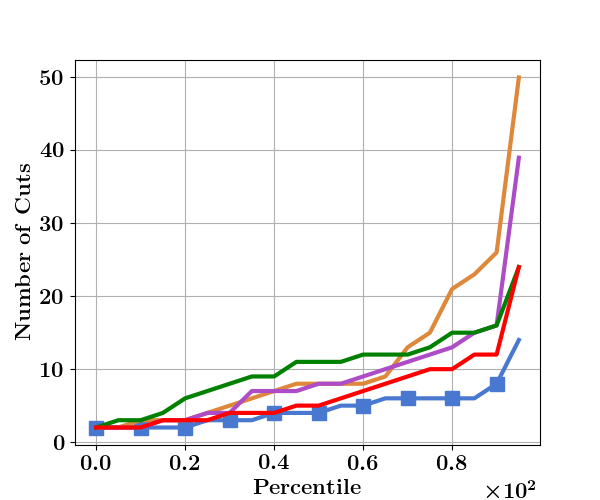}}
\caption{\small{Percentage of cuts meeting the designed criteria and number of cuts on Knapsack problems. We train the RL agent on $80$ instances. All baselines are tested on $20$ instances. As seen above, \gls{RL} produces consistently more 'high-quality' cuts.}}
\label{figure:classificationscores}
\end{figure*}

\paragraph{Experiment \#3: Generalization.}

In  Figure \ref{figure:objective_medium},\sacomment{I removed the line from the table, I think it is not required, the figure shows it follows very closely the RL}\removed{and Table \ref{table:optimalitygap_medium},} 
we also demonstrate the ability of the RL agent to generalize across different sizes of the \gls{IP} instances.
 This is illustrated through the extremely competitive performance of the RL/10X agent, which is trained on 10X smaller size instances than the test instances. 
 (Exact sizes used in the training of RL/10X agent were were $10\times 10, 32 \times 22, 10 \times 20, 20 \times 10$, respectively, for the four types of \gls{IP} problems.) 
 Furthermore, we test generalizability across \gls{IP} classes by training an RL agent on small sized instances of the {\it packing problem}, and applying it to add cuts to $10X$ larger instances of the {\it maximum-cut problem}. The latter, being a graph optimization problem, has  intuitively a very different structure from the former. Figure~\ref{figure:crossgeneralization} shows 
that the RL/10X agent trained on packing (yellow curve) 
achieve a performance on larger maximum-cut instances that is comparable to the performance of agent trained on the latter class (blue curve).


\paragraph{Experiment \#4: Impact on the efficiency of \gls{bnc}.}
In practice, cutting planes alone are not sufficient to solve large problems. 
In state-of-the-art solvers, the iterative addition of cutting planes is alternated with a branching procedure, leading to Branch-and-Cut (B\&C). 
To demonstrate the full potential of \gls{RL}, we implement a comprehensive B\&C procedure but without all the additional heuristics that appear in the standard solvers. Our B\&C procedure has two hyper-parameters: number of child nodes (suproblems) to expand $N_{\text{exp}}$ and number of cutting planes added to each node $N_{\text{cuts}}$. 
In addition, \gls{bnc} is determined by the implementation of the {\it Branching Rule}, {\it Priority Queue} and {\it Termination Condition}. Further details are in the appendix.

Figure \ref{figure:numberofbb} gives percentile plots for the number of child nodes (suproblems) $N_{\text{exp}}$ until termination of \gls{bnc}. Here, $N_{\text{cuts}}=10$ cuts were added to each node, using either RL or one of the baseline heuristics. We also include as a comparator, the \gls{bnc} method without any cuts, i.e., the branch and bound method.  \added{The trained RL agent and the test instances used here are same as those in Table \ref{table:optimalitygap_medium} and Figure \ref{figure:objective_medium}.}  Perhaps surprisingly, the RL agent, though not designed to be used in combination with branching, shows substantial improvements in the efficiency of \gls{bnc}. 
\added{In the appendix, we have also included  experimental results showing improvements for the   instances used in Table \ref{table:optimalitygap_largescale} and Figure \ref{figure:objective_largescale}}.

\paragraph{Experiment \#5: Interpretability of cuts.} A \emph{knapsack} problem is a binary packing problem with only one constraint. Although simple to state, these problems are NP-Hard, and have been a testbed for many algorithmic techniques, see e.g.~the books~\cite{kellerer2003knapsack,martello1990knapsack}. A prominent class of valid inequalities for knapsack is that of \emph{cover inequalities}, that can be strengthened through the classical \emph{lifting operation} (see the appendix for definitions). Those inequalities are well-studied in theory and also known to be effective in practice, see e.g.~\cite{crowder1983solving,conforti2014integer,kellerer2003knapsack}. Our last set of experiments gives a ``reinforcement learning validation'' of those cuts. We show in fact that \gls{RL}, with the same reward scheme as in Experiment \#2, produces many more cuts that ``almost look like'' lifted cover inequalities than the baselines. More precisely, we define three increasingly looser criteria for deciding when a cut is ``close'' to a lifted cover inequality (the plurality of criteria is due to the fact that lifted cover inequalities can be strengthened in many ways). We then check which percentage of the inequalities produced by the \gls{RL} (resp.~the other baselines) satisfy each of these criteria. This is reported in the first three figures in  Figure~\ref{figure:classificationscores}, together with the number of cuts added before the optimal solution is reached (rightmost figure in  Figure~\ref{figure:classificationscores}). More details on the experiments and a description of the three criteria are reported in the appendix.
These experiments suggest that our approach could be useful to aid researchers in the discovery of strong family of cuts for \gls{IP}s, and provide yet another empirical evaluation of known ones.

\added{\paragraph{Runtime.} 
A legitimate question is whether the improvement provided by the RL agent in terms of solution accuracy comes at the cost of a large runtime. The training time for RL can indeed be significant, especially when trained on large instances. However, there is no way to compare that with the human-designed heuristics. In testing, we observe no significant differences in time required by an RL policy to choose a cut vs.~time taken to execute a heuristic rule.  
We detail the runtime comparison in the appendix.}

\section{Conclusions} 
\added{We presented a deep RL approach to automatically learn an effective cutting plane strategy for solving IP instances. The RL algorithm learns by trying to solve a pool of (randomly generated) training instances again and again, without having access to any solved instances.} The variety of tasks across which the RL agent is demonstrated to generalize without being trained for, provides evidence that it is able to learn an intelligent algorithm for selecting cutting planes. We believe our empirical results are a convincing step forward towards the integration of  ML techniques in IP solvers. This may lead to a functional answer to the ``Hamletic question Branch-and-cut designers often have to face: to cut or not to cut?''  \citep{dey2018theoretical}.

\paragraph{Acknowledgements.}
Author Shipra Agrawal acknowledges support from an Amazon Faculty Research Award.

\bibliographystyle{apa}
\bibliography{your_bib_file.bib}

\begin{thebibliography}{}

\bibitem[\protect\astroncite{Balas et~al.}{1993}]{balas1993lift}
Balas, E., Ceria, S., and Cornu{\'e}jols, G. (1993).
\newblock A lift-and-project cutting plane algorithm for mixed 0--1 programs.
\newblock {\em Mathematical programming}, 58(1-3):295--324.

\bibitem[\protect\astroncite{Balcan et~al.}{2018}]{balcan2018learning}
Balcan, M.-F., Dick, T., Sandholm, T., and Vitercik, E. (2018).
\newblock Learning to branch.
\newblock {\em arXiv preprint arXiv:1803.10150}.

\bibitem[\protect\astroncite{Bello et~al.}{2017}]{bello2017neural}
Bello, I., Pham, H., Le, Q.~V., Norouzi, M., and Bengio, S. (2017).
\newblock Neural combinatorial optimization.

\bibitem[\protect\astroncite{Bengio et~al.}{2018}]{bengio2018machine}
Bengio, Y., Lodi, A., and Prouvost, A. (2018).
\newblock Machine learning for combinatorial optimization: a methodological
  tour d'horizon.
\newblock {\em arXiv preprint arXiv:1811.06128}.

\bibitem[\protect\astroncite{Bixby}{2017}]{bixby}
Bixby, B. (2017).
\newblock Optimization: past, present, future.
\newblock Plenary talk at INFORMS Annual Meeting.

\bibitem[\protect\astroncite{Conforti et~al.}{2014}]{conforti2014integer}
Conforti, M., Cornu{\'e}jols, G., and Zambelli, G. (2014).
\newblock {\em Integer programming}, volume 271.
\newblock Springer.

\bibitem[\protect\astroncite{Cornu{\'{e}}jols et~al.}{2013}]{CornuejolsMargot}
Cornu{\'{e}}jols, G., Margot, F., and Nannicini, G. (2013).
\newblock On the safety of gomory cut generators.
\newblock {\em Math. Program. Comput.}, 5(4):345--395.

\bibitem[\protect\astroncite{Crowder et~al.}{1983}]{crowder1983solving}
Crowder, H., Johnson, E.~L., and Padberg, M. (1983).
\newblock Solving large-scale zero-one linear programming problems.
\newblock {\em Operations Research}, 31(5):803--834.

\bibitem[\protect\astroncite{Dai et~al.}{2017}]{dai2017learning}
Dai, H., Khalil, E.~B., Zhang, Y., Dilkina, B., and Song, L. (2017).
\newblock Learning combinatorial optimization algorithms over graphs.
\newblock {\em arXiv preprint arXiv:1704.01665}.

\bibitem[\protect\astroncite{Dantzig et~al.}{1954}]{dantzig1954solution}
Dantzig, G., Fulkerson, R., and Johnson, S. (1954).
\newblock Solution of a large-scale traveling-salesman problem.
\newblock {\em Journal of the operations research society of America},
  2(4):393--410.

\bibitem[\protect\astroncite{Dey and Molinaro}{2018}]{dey2018theoretical}
Dey, S.~S. and Molinaro, M. (2018).
\newblock Theoretical challenges towards cutting-plane selection.
\newblock {\em Mathematical Programming}, 170(1):237--266.

\bibitem[\protect\astroncite{Espeholt et~al.}{2018}]{espeholt2018impala}
Espeholt, L., Soyer, H., Munos, R., Simonyan, K., Mnih, V., Ward, T., Doron,
  Y., Firoiu, V., Harley, T., Dunning, I., et~al. (2018).
\newblock Impala: Scalable distributed deep-rl with importance weighted
  actor-learner architectures.
\newblock {\em arXiv preprint arXiv:1802.01561}.

\bibitem[\protect\astroncite{Gomory}{1960}]{gomory1960algorithm}
Gomory, R. (1960).
\newblock An algorithm for the mixed integer problem.
\newblock Technical report, RAND CORP SANTA MONICA CA.

\bibitem[\protect\astroncite{Gurobi~Optimization}{2015}]{gurobi2015gurobi}
Gurobi~Optimization, I. (2015).
\newblock Gurobi optimizer reference manual.
\newblock {\em URL http://www. gurobi. com}.

\bibitem[\protect\astroncite{Hochreiter and
  Schmidhuber}{1997}]{hochreiter1997long}
Hochreiter, S. and Schmidhuber, J. (1997).
\newblock Long short-term memory.
\newblock {\em Neural computation}, 9(8):1735--1780.

\bibitem[\protect\astroncite{Kapturowski
  et~al.}{2018}]{kapturowski2018recurrent}
Kapturowski, S., Ostrovski, G., Quan, J., Munos, R., and Dabney, W. (2018).
\newblock Recurrent experience replay in distributed reinforcement learning.

\bibitem[\protect\astroncite{Kellerer et~al.}{2003}]{kellerer2003knapsack}
Kellerer, H., Pferschy, U., and Pisinger, D. (2003).
\newblock Knapsack problems. 2004.

\bibitem[\protect\astroncite{Khalil et~al.}{2017}]{khalil2017learning}
Khalil, E., Dai, H., Zhang, Y., Dilkina, B., and Song, L. (2017).
\newblock Learning combinatorial optimization algorithms over graphs.
\newblock In {\em Advances in Neural Information Processing Systems}, pages
  6348--6358.

\bibitem[\protect\astroncite{Khalil et~al.}{2016}]{khalil2016learning}
Khalil, E.~B., Le~Bodic, P., Song, L., Nemhauser, G.~L., and Dilkina, B.~N.
  (2016).
\newblock Learning to branch in mixed integer programming.
\newblock In {\em AAAI}, pages 724--731.

\bibitem[\protect\astroncite{Kingma and Ba}{2014}]{kingma2014adam}
Kingma, D.~P. and Ba, J. (2014).
\newblock Adam: A method for stochastic optimization.
\newblock {\em arXiv preprint arXiv:1412.6980}.

\bibitem[\protect\astroncite{Kool and Welling}{2018}]{kool2018attention}
Kool, W. and Welling, M. (2018).
\newblock Attention solves your tsp.
\newblock {\em arXiv preprint arXiv:1803.08475}.

\bibitem[\protect\astroncite{Li et~al.}{2018}]{li2018combinatorial}
Li, Z., Chen, Q., and Koltun, V. (2018).
\newblock Combinatorial optimization with graph convolutional networks and
  guided tree search.
\newblock In {\em Advances in Neural Information Processing Systems}, pages
  539--548.

\bibitem[\protect\astroncite{Martello and Toth}{1990}]{martello1990knapsack}
Martello, S. and Toth, P. (1990).
\newblock Knapsack problems: algorithms and computer implementations.
\newblock {\em Wiley-Interscience series in discrete mathematics and optimiza
  tion}.

\bibitem[\protect\astroncite{Mnih et~al.}{2016}]{mnih2016asynchronous}
Mnih, V., Badia, A.~P., Mirza, M., Graves, A., Lillicrap, T., Harley, T.,
  Silver, D., and Kavukcuoglu, K. (2016).
\newblock Asynchronous methods for deep reinforcement learning.
\newblock In {\em International conference on machine learning}, pages
  1928--1937.

\bibitem[\protect\astroncite{Nowak et~al.}{2017}]{nowak2017note}
Nowak, A., Villar, S., Bandeira, A.~S., and Bruna, J. (2017).
\newblock A note on learning algorithms for quadratic assignment with graph
  neural networks.
\newblock {\em arXiv preprint arXiv:1706.07450}.

\bibitem[\protect\astroncite{Pochet and Wolsey}{2006}]{pochet2006production}
Pochet, Y. and Wolsey, L.~A. (2006).
\newblock {\em Production planning by mixed integer programming}.
\newblock Springer Science \& Business Media.

\bibitem[\protect\astroncite{Rothvo{\ss} and
  Sanit{\`a}}{2017}]{rothvoss-sanita}
Rothvo{\ss}, T. and Sanit{\`a}, L. (2017).
\newblock 0/1 polytopes with quadratic chv{\'a}tal rank.
\newblock {\em Operations Research}, 65(1):212--220.

\bibitem[\protect\astroncite{Salimans et~al.}{2017}]{salimans2017evolution}
Salimans, T., Ho, J., Chen, X., Sidor, S., and Sutskever, I. (2017).
\newblock Evolution strategies as a scalable alternative to reinforcement
  learning.
\newblock {\em arXiv preprint arXiv:1703.03864}.

\bibitem[\protect\astroncite{Sutskever et~al.}{2014}]{sutskever2014sequence}
Sutskever, I., Vinyals, O., and Le, Q.~V. (2014).
\newblock Sequence to sequence learning with neural networks.
\newblock In {\em Advances in neural information processing systems}, pages
  3104--3112.

\bibitem[\protect\astroncite{Tokui et~al.}{2015}]{tokui2015chainer}
Tokui, S., Oono, K., Hido, S., and Clayton, J. (2015).
\newblock Chainer: a next-generation open source framework for deep learning.
\newblock In {\em Proceedings of workshop on machine learning systems
  (LearningSys) in the twenty-ninth annual conference on neural information
  processing systems (NIPS)}, volume~5, pages 1--6.

\bibitem[\protect\astroncite{Vaswani et~al.}{2017}]{vaswani2017attention}
Vaswani, A., Shazeer, N., Parmar, N., Uszkoreit, J., Jones, L., Gomez, A.~N.,
  Kaiser, {\L}., and Polosukhin, I. (2017).
\newblock Attention is all you need.
\newblock In {\em Advances in neural information processing systems}, pages
  5998--6008.

\bibitem[\protect\astroncite{Vinyals et~al.}{2015}]{vinyals2015pointer}
Vinyals, O., Fortunato, M., and Jaitly, N. (2015).
\newblock Pointer networks.
\newblock In {\em Advances in Neural Information Processing Systems}, pages
  2692--2700.

\bibitem[\protect\astroncite{Wesselmann and
  Stuhl}{2012}]{wesselmann2012implementing}
Wesselmann, F. and Stuhl, U. (2012).
\newblock Implementing cutting plane management and selection techniques.
\newblock Technical report, Tech. rep., University of Paderborn.

\end{thebibliography}
\newpage
\appendix
\onecolumn{}
\section{Experiment Details}
\subsection{Projection into the original variable space}
In the following we look at only the first iteration of the cutting plane procedure, and we drop the iteration index $t$. Recall the LP relaxation of the original IP problem (\ref{eq:ip}), where $A \in \mathbb{Q}^{m\times n},b\in \mathbb{Q}^m$:
\begin{equation}
    \left\{
                \begin{array}{ll}
                  \min c^T x\\
                  Ax \leq b\\
                  x \geq 0.
                \end{array}
              \right. 
\nonumber
\end{equation}
When a simplex algorithm solves the LP, the original LP is first converted to a standard form where all inequalities are transformed into equalities by introducing slack variables. 
\begin{equation}
    \left\{
                \begin{array}{ll}
                  \min c^T x \\
                  Ax + \mathbb{I}s = b\\
                  x \geq 0, s\geq 0,
                \end{array}
              \right. 
\label{eq:slacklp}
\end{equation}
where $\mathbb{I}$ is an identity matrix and $s$ is the set of slack variables. The simplex method carries out iteratively operations on the tableau formed by $[A,\mathbb{I}],b$ and $c$. At convergence, the simplex method returns a final optimal tableau. We generate a Gomory's cut using the row of the tableau that corresponds to a fractional variable of the optimal solution $x_{\text{LP}}^\ast$. This will in general create a cutting plane of the following form
\begin{align}
    e^T x + r^T s \leq d
    \label{eq:newcut}
\end{align}
where $e,x\in \mathbb{R}^n, r,s\in \mathbb{R}^m$ and $d \in \mathbb{R}$. Though this cutting plane involves slack variables, we can get rid of the slack variables by multiplying both sides of the linear constraints in (\ref{eq:slacklp}) by $r$
\begin{align}
    r^T Ax + r^T s = r^T b
\end{align}
and subtract the new cutting plane (\ref{eq:newcut}) by the above. This leads to an equivalent cutting plane
\begin{align}
    (e^T - r^T A) x \leq  d - r^T b .
\end{align}
Note that this cutting plane only contains variables in the original variable space. For a downstream neural network that takes in the parameters of the cutting planes as inputs, we find it helpful to remove such slack variables. Slack variables do not contribute to new information regarding the polytope and we can also parameterize a network with a smaller number of parameters.

\subsection{Integer programming formulations of benchmark problems}
A wide range of benchmark instances can be cast into special cases of IP problems. We provide their specific formulations below. For simplicity, we only provide their general IP formulations (with $\leq,\geq,=$ constraints). It is always possible to convert original formulations into the standard formulation (\ref{eq:ip}) with properly chosen $A,b,c,x$. Some problems are formulated within a graph $G = (V,E)$ with nodes $v \in V$ and edges $(v,u) \in E$.

 Their formulations are as follows:

\paragraph{Max Cut.} We have one variable per edge $y_{u,v},(u,v)\in E$ and one variable per node $x_u,u\in V$. Let $w_{u,v} \geq 0$ be a set of non-negative weights per edge.
\begin{equation}
    \left\{
                \begin{array}{ll}
                  \max \sum_{(u,v) \in E} w_{uv}y_{uv}\\
                  y_{uv} \leq x_u + x_v, \forall (u,v) \in E\\
                  y_{uv} \leq 2-x_u-x_v, \forall (u,v) \in E\\
                  0 \leq x,y  \leq 1 \\
                  x_u, y_{uv} \in \mathbb{Z} \quad  \forall u \in V, (u,v) \in E.
                \end{array}
              \right. 
\label{eq:mwm}
\end{equation}

In our experiments the graphs are randomly generated. To be specific, we specify a vertex size $|V|$ and edge size $|E|$. We then sample $|E|$ edges from all the possible $|V|\cdot(|V|-1) / 2$ edges to form the final graph. The weights $w_{uv}$ are uniformly sampled as an integer from $0$ to $10$. When generating the instances, we sample graphs such that $|V|,|E|$ are of a particular size. For example, for middle size problem we set $|V|=7, |E|=20$.

\paragraph{Packing.} The packing problem takes the generic form of (\ref{eq:ip}) while requiring that all the coefficients of $A,b,c$ be non-negative, in order to enforce proper resource constraints.

Here the constraint coefficients $a_{ij}$ for the $j$th variable and $i$th constraint is sampled as an integer uniformly from $0$ and $5$. Then the RHS coefficient $b_i$ is sampled from $9n$ to $10n$ uniformly as an integer where $n$ is the number of variables. Each component of $c_j$ is uniformly sampled as an integer from $1$ to $10$.

\paragraph{Binary Packing.} Binary packing augments the original packing problem by a set of binary constraints on each variable $x_i \leq 1$. 

Here the constraint coefficients $a_{ij}$ for the $j$th variable and $i$th constraint is sampled as an integer uniformly from $5$ and $30$. Then the RHS coefficient $b_i$ is sampled from $10n$ to $20n$ uniformly as an integer where $n$ is the number of variables. Each component of $c_j$ is uniformly sampled as an integer from $1$ to $10$.

\paragraph{Production Planning.}
Consider a production planning problem \citep{pochet2006production} with time horizon $T$. The decision variables are production $x_i, 1\leq i \leq T$, along with by produce / not produce variables $y_i,1\leq i\leq T$ and storage variables $s_i,0\leq i\leq T$. Costs $p_i^\prime, h_i^\prime, q_i$ and demands $d_i$ are given as problem parameters. The LP formulation is as follows
\begin{equation}
    \left\{
                \begin{array}{ll}
                  \min \sum_{i=1}^T p_i^\prime x_i + \sum_{i=0}^T h_i^\prime s_i + \sum_{i=0}^T q_i y_i\\
                  s_{i-1} + x_i = d_i + s_i, \forall 1\leq i\leq T\\
                  x_i \leq M y_i, \forall 1\leq i\leq T\\
                  s\geq 0, x \geq 0, 0\leq y\leq 1\\
                  s_0 = s_0^\ast, s_T = s_T^\ast\\
                  x,s,y \in \mathbb{Z}^T,
                \end{array}
              \right. 
\label{eq:maxcut}
\end{equation}
where $M$ is a positive large number and $s_0^\ast,s_T^\ast$ are also given.

The instance parameters are the initial storage $s_0^\ast = 0$, final storage $s_T^\ast = 20$ and big $M = 100$. The revenue parameter $p_i^\prime,h_i^\prime,q_i$ are generated uniformly random as integers from $1$ to $10$.

\paragraph{Size of IP formulations.}
In our results, we describe the sizes of the \gls{IP} instances as $n\times m$ where $n$ is the number of columns and $m$ is the number of rows of the constraint matrix $A$ from the LR of~\eqref{eq:ip}. For a packing problem with $n$ items and $m$ resource constraints, the \gls{IP} formulation has $n$ variables and $m$ constraints; for planning with period $K$, $n=3K+1$, $m=4K+1$; for binary packing, there are $n$ extra binary constraints compared to the packing problem; for max-cut, the problem is defined on a graph with a vertex set $V$ and an edge set $V$, and its \gls{IP} formulation consists of $n=|V|+ | E|$ variables and $m=3| E|+ |V|$ constraints.

\subsection{Criteria for selecting Gomory cuts}

 Recall that $I_t$ is the number of candidate Gomory cuts available in round $t$, and $i_t$ denotes the index of cut chosen by a given baseline. The baseline heuristics we use are the following: 
\begin{itemize}
   \item Random. One cut $i_t \sim \text{Uniform}\{1,2...I_t\}$ is chosen uniformly at random from all the candidate cuts. 
    \item Max Violation (MV). Let $x_B^\ast(t)$ be the basic feasible solution of the curent LP relaxation. MV selects the cut that corresponds to the most fractional component, i.e. $i_t = \arg\max \{|[x_B^\ast(t)]_i - \text{round}([x_B^\ast(t)]_i)|\}$.
   \item Max Normalized Violation (MNV). Recall that $\tilde{A}$ denotes the optimal tableau obtained by the simplex algorithm upon convergence. Let $\tilde{A}_i$ be the $i$th row of $\tilde{A}$. Then, MNV selects cut $i_t = \arg\max \{|[x_B^\ast(t)]_i - \text{round}([x_B^\ast(t)]_i)| / \Vert \tilde{A}_i \Vert\} $.
   \item Lexicographic (LE): Add the cutting plane with the least index, i.e. $i_t = \arg\min\{i, [x_B^\ast(t)]_i\text{ is fractional}\}$.
\end{itemize}

The first three rules are common in the \gls{IP} literature, see e.g.~\citep{wesselmann2012implementing}, while the fourth is the original rule used by Gomory to prove the convergence of his method~\citep{gomory1960algorithm}.

\subsection{Hyper-parameters}
\paragraph{Policy architecture.}
The policy network is implemented with Chainer \citep{tokui2015chainer}. The attention embedding $F_\theta$ is a 2-layer neural network with $64$ units per layer and $\text{tanh}$ activation. The LSTM network encodes variable sized inputs into hidden vector with dimension $10$. 

During a forward pass, a LSTM $+$ Attention policy will take the instance, carry out embedding into a $n$-d vector and then apply attention. Such architecture allows for generalization to variable sized instances (different number of variables). We apply such architecture in the generalization part of the experiments.

On the other hand, a policy network can also consist of a single attention network. This policy can only process \gls{IP} instances of a fixed size (fixed number of varibles) and cannot generalize to other sizes. We apply such architecture in the \gls{igc} part of the experiments.

\paragraph{\gls{ES} optimization.} Across all experiments, we apply Adam optimizer \citep{kingma2014adam} with learning rate $\alpha = 0.01$ to optimize the policy network. The perturbation standard deviation $\sigma$ is selected from  $\{0.002,0.02,0.2\}$. By default, we apply $N=10$ perturbations to construct the policy gradient for each iteration, though we find that $N=1$ could also work as well. For all problem types except planning, we find that $\sigma=0.2$ generally works properly except for planning, where we apply $\sigma=0.02$ and generate $N=5$ trajectory per instance per iteration. Empirically, we observe that the training is stable for both policy architectures and the training performance converges in $\leq 500$ weight updates.

\paragraph{Distributed setup.} For training, we use a Linux machine with 60 virtual CPUs. To fully utilize the compute power of the machine, the trajectory collection is distributed across multiple workers, which run in parallel.

\section{Branch-and-Cut Details}
As mentioned in the introduction, Branch-and-Cut (B\&C) is an algorithmic procedure used for solving IP problems. The choice of which variable to branch on, as well as which node of the branching tree to explore next, is the subject of much research. In our experiments, we implemented a B\&C with very simple rules, as explained below. This is motivated by the fact that our goal is to evaluate the quality of the cutting planes added by the RL rather than obtaining a fast B\&C method. Hence, sophisticated and computationally expensive branching rules could have overshadowed the impact of cutting planes. Instead, simple rules (applied both to the RL and to the other techniques) highlight the impact of cutting planes for this important downstream application. 

We list next several critical elements of our implementation of B\&C.

\paragraph{Branching rule.} At each node, we branch on the most fractional variable of the corresponding LP optimal solution ($0.5$ being the most fractional).
\paragraph{Priority queue.} 
We adopt a FIFO queue (Breath first search). 
FIFO queue allows the B\&C procedure to improve the lower bound. 
\paragraph{Termination condition.} Let $z_0 = c^T x_{\text{LP}}^\ast(0)$ be the objective of the initial LP relaxation. As B\&C proceeds, the procedure finds an increasing set of feasible integer solutions $\mathcal{X}_F$, and an upper bound on the optimal objective $z^\ast = c^T x_{\text{IP}}^\ast$ is $z_{\text{upper}} = \min_{x\in \mathcal{X}_F} c^T x$. Hence, $z_{\text{upper}}$ monotonically decreases.

Along with B\&C, cutting planes can iteratively improve the lower bound $z_{\text{lower}}$ of the optimal objective $z^\ast$. Let $z_i$ be the objective of the LP solution at node $i$ and denote $\mathcal{N}$ as the set of unpruned nodes with unexpanded child nodes. The lower bound is computed as 
$z_{\text{lower}} = \min_{i\in \mathcal{N}} z_i$ and monotonically increases as the B\&C procedure proceeds.

This produces a ratio statistic
\begin{align}
    r = \frac{z_{\text{upper}}-z_{\text{lower}} }{z_{\text{upper}}-z^*_{\text{LP} }} > 0\nonumber
\end{align}
Note that since $z_{\text{lower}}\geq z^*_{\text{LP}}$,  $z_{\text{lower}}$ monotonically increases, and $z_{\text{upper}}$ monotonically decreases, $r$ monotonically decreases. The B\&C terminates when $r$ is below some threshold which we set to be $0.0001$.

\section{Test Time Considerations}
\paragraph{Stopping criterion.}
Though at training time we guide the agent to generate aggressive cuts that tighten the \gls{LP} relaxation as much as possible, the agent can exploit the defects in the simulation environment - numerical errors, and generate invalid cuts which cut off the optimal solution.

This is undesirable in practice. In certain cases at test time, when we execute the trained policy, we adopt a \emph{stopping criterion} which automatically determines if the agent should stop adding cuts, in order to prevent from invalid cuts. In particular, at each iteration let $r_t = |c^Tx_{\text{LP}^\ast(t)} - c^T x_{\text{LP}^\ast(t+1)}|$ be the objective gap achieved by adding the most recent cut. We maintain a cumulative ratio statistics such that
\begin{align}
    s_t = \frac{r_t}{\sum_{t^\prime \leq t}r_t}.\nonumber
\end{align}
We terminate the cutting plane procedure once the average $s_t$ over a fixed window of size $H$ is lower than certain threshold $\eta$. In practice, we set $H = 5,\eta = 0.001$ and find this work effectively for all problems, eliminating all the numerical errors observed in reported tasks. Intuitively, this approach dictates that we terminate the cutting plane procedure once the newly added cuts do not generate significant improvements for a period of $H$ steps.

To analyze the effect of $\eta$ and $H$, we note that when $H$ is too small or $\eta$ is too large, we have very conservative cutting plane procedure. On the other hand when $H$ is large while $\eta$ is small, the cutting plane procedure becomes more aggressive.
 
\paragraph{Greedy action.} The policy network defines a stochastic policy, i.e. a categorical distribution over candidate cuts. At test time, we find taking the greedy action $i^\ast = \arg\max p_i$ to be more effective in certain cases, where $p_i$ is the categorical distribution over candidate cuts. The justification for this practice is that: the \gls{ES} optimization procedure can be interpreted as searching for a parameter $\theta$ such that the induced distribution over trajectories has large concentration on those high return trajectories. Given a trained model, to decode the \emph{most likely} trajectory of horizon $T$ generated by the policy, we need to run a full tree search of depth $T$, which is infeasible in practice. Taking the greedy action is equivalent to applying a greedy strategy in decoding the most likely trajectory.

This approach is highly related to beam search in sequence modeling \citep{sutskever2014sequence} where the goal is to decode the prediction that the model assigns the most likelihood to. The greedy action selection above corresponds to a beam search with $1$-step lookahead.

\section{Details on the Interpretation of Cutts}
One interesting aspect of studying the \gls{RL} approach to generating cuts, is to investigate if we can interpret cuts generated by \gls{RL}. For a particular class of \gls{IP} problems, certain cuts might be considered as generally 'better' than other cuts. For example, these cuts might be more effective in terms of closing the objective gap, according to domain knowledge studied in prior literature. Ideally, we would like to find out what \gls{RL} has learned, whether it has learned to select these more 'effective' cuts with features identified by prior works. Here, we focus on \emph{Knapsack problems}.


\paragraph{Problem instances.} Consider the knapsack problems
\begin{equation}
    \left\{
                \begin{array}{ll}
                  \max \sum_{i}^n c_i x_i\\
                  \sum_{i=1}^n a_i x_i \leq \beta := \sum_{i=1}^n a_i / 2\\
                  x_i\in \{0,1\}, \\
                \end{array}
              \right. 
\label{eq:knapsackoneconstraint}
\end{equation}
where $a_i$ are generated independently and uniformly in $[1, 30]$ as integers, and the $c_i$ are generated independently and uniformly in $[1, 10]$. We consider $n=10$ in our experiments. Knapsack problems are fundamental in \gls{IP}, see e.g.~\cite{kellerer2003knapsack}. The intuition of the problem is that we attempt to pack as many items as possible into the knapsack, as to maximize the profit of the selected items. Polytopes as~\eqref{eq:knapsackoneconstraint} are also used to prove strong (i.e., quadratic) lower bounds on the Chv\'atal-Gomory rank of polytopes with $0/1$ vertices~\cite{rothvoss-sanita}.  
\paragraph{Evaluation scores.} For knapsack problems, one effective class of cuts is given by \emph{cover inequalities}, and their strengthening through \emph{lifting}~\cite{conforti2014integer,kellerer2003knapsack}. The cover inequality associated to a set $S\subseteq \{1,\dots,n\}$ with $\sum_{i \in S} a_i >\beta$ and $|S|=k$ is given by  $$\sum_{i \in S} x_i \leq k-1.$$ Note that cover inequalities are valid for~\eqref{eq:knapsackoneconstraint}. The inequality can be strengthened (while maintaining validity) by replacing the $0$ coefficients of variables $x_i$ for $i \in \{1,\dots,n\}\setminus S$ with appropriate positive coefficients, leading to the \emph{lifted cover inequality} below:
\begin{equation}\label{eq:lifted-cover}
    \sum_{i \in S} x_i + \sum_{ \notin S} \alpha_i x_i \leq k-1.
\end{equation}
with all $\alpha_i \geq 0$. There are in general exponentially many ways to generate lifted cover inequalities from a single cover inequality. In practice, further strengthenings are possible, for instance, by perturbing the right-hand side or the coefficients of $x_i$ for $i \in S$. We provide three criteria for identifying (strengthening of) lifted cover inequalities, each capturing certain features of the inequalities (below, RHS denotes the right-hand side of a given inequality). 
\begin{itemize}
\item[1.] There exists an integer $p$ such that (1) the RHS is an integer multiple of $p$ and (2) $p$ times (number of variable with coefficient exactly $p$) $>$ RHS.
\end{itemize}
Criterion 1 is satisfied by all lifted cover inequalities as in~\eqref{eq:lifted-cover}. The scaling by $p$ is due to the fact that an inequality may be scaled by a positive factor, without changing the set of points satisfying it. \yfcomment{Do we say somewhere that all inequalities produced have integer coefficients?} 

\begin{itemize}

\item[2.] There exists an integer $p$ such that (1) holds and (2') $p$ times (number of variables with coefficients between $p$ and $p+2$) $>$ RHS.

\item[3.] There exists an integer $p$ such that (1) holds and (2'') p times (number of variables with coefficients at least $p$) $>$ RHS.
\end{itemize}

A lifted cover inequality can often by strengthened by increasing the coefficients of variables in $S$, after the lifting has been performed. We capture this by criteria 2 and 3 above, where $2$ is a stricter criterion, as we only allow those variables to have their coefficients increased by a small amount.


For each cut $c_j$ generated by the baseline (e.g.~\gls{RL}), we evaluate if this cut satisfies the aforementioned conditions. For one particular condition, if satisfied, the cut is given a score $s(c_j) = 1$ or else $s(c_j) = 0$. On any particular instance, the overall score is computed as an average across the $m$ cuts that are generated to solve the problem with the cutting plane method:
\begin{align*}
\frac{1}{m}\sum_{j=1}^m s(c_j) \in [0,1].
\end{align*}

\paragraph{Evaluation setup.} We train a \gls{RL} agent on $100$ knapsack instances and evaluate the scores on another independently generated set of $20$ instances. Please see the main text for the evaluation results.




\section{Additional results on Large-scale Instances}

\begin{figure}[h!]
\centering
\subfigure[Packing (25\% IGC, 1000 nodes)]{\includegraphics[width=.48\linewidth]{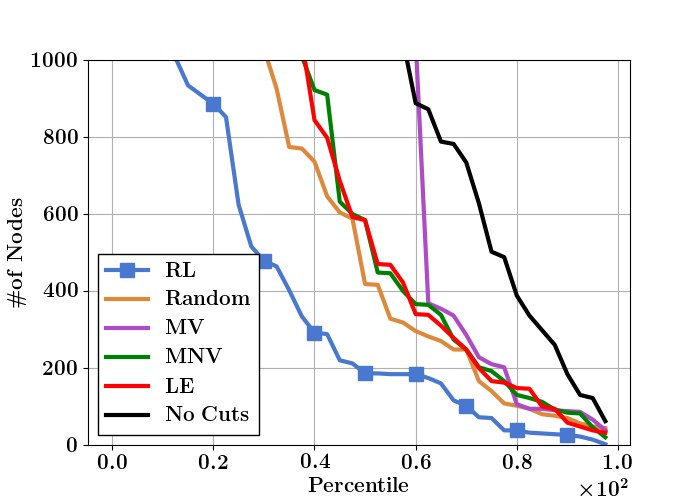}}
\subfigure[Planning (95\% IGC, 1000 nodes)]{\includegraphics[width=.48\linewidth]{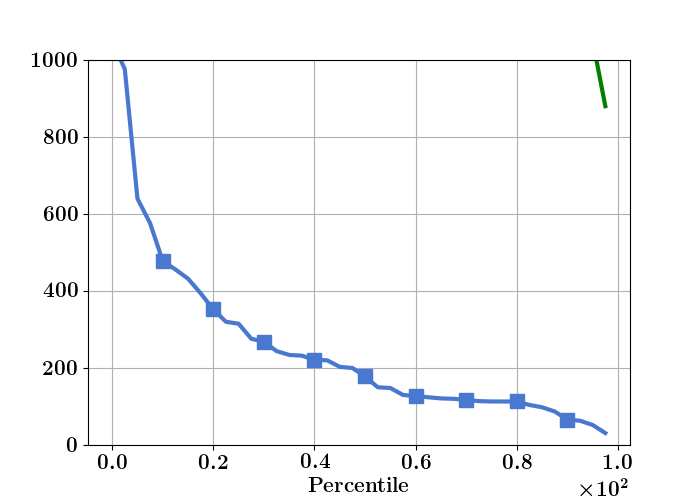}}
\subfigure[Binary Packing (95\% IGC, 1000 nodes) ]{\includegraphics[width=.48\linewidth]{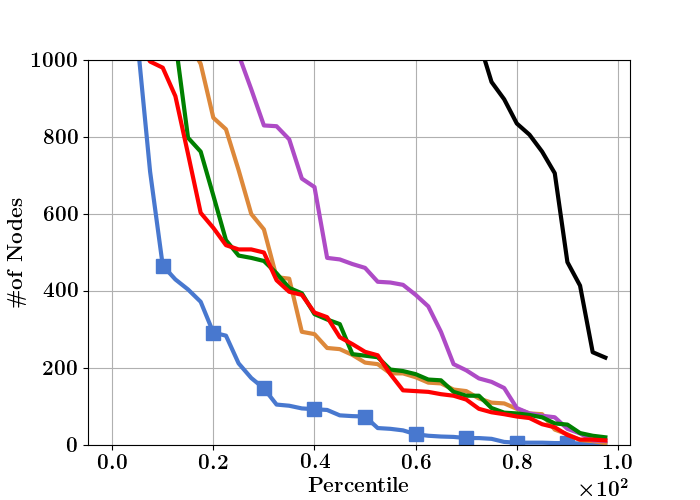}}
\subfigure[Max Cut (95\% IGC, 1000 nodes)]{\includegraphics[width=.48\linewidth]{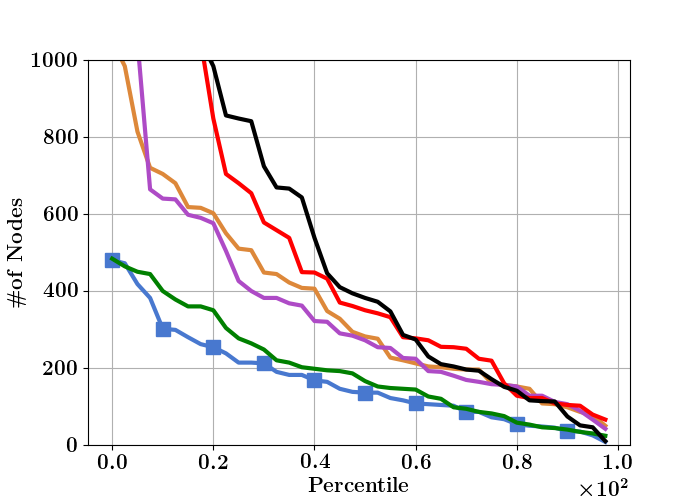}}
\caption{\small{Percentile plots of number of B\&C nodes expanded for large-scale instances. The same setup as Figure \ref{figure:numberofbb} but for even larger instances.}}
\label{figure:numberofbb_largescale}
\end{figure}

We provide additional results on large-scale instances in Figure \ref{figure:numberofbb_largescale}, in the context of \gls{bnc}. Experimental setups and details are similar to those of the main text: we set the threshold limit to be $1000$ nodes for all problem classes. The results show the percentile plots of the number of nodes required to achieve a certain level of IGC during the \gls{bnc} with the use of cutting plane heuristics, where the percentile is calculated across instances. Baseline results for each baseline are shown via curves in different colors. When certain curves do not show up in the plot, this implies that these heuristics do not achieve the specified level of IGC within the node budgets. The IGC level is set to be $95$\% as in the main text, except for the random packing problem where it is set to be $25$\%. 

The IGC of the random packing is set at a relatively low level because random packing problems are significantly more difficult to solve when instances are large-scaled. This is consistent with the observations in the main text.

Overall, we find that the performance of \gls{RL} agent significantly exceeds that of the other baseline heuristics. For example, on the planning problem, other heuristics barely achieve the IGC within the node budgets. There are also cases where \gls{RL} does similarly to certain heuristics, such as to MNV on the Max Cut problems.

\begin{table}[t]
\caption{IGC in \gls{bnc} with large-scale instances. We adopt the same setup as Table \ref{table:optimalitygap_medium}}
\begin{center}
\begin{small}
\begin{sc}
\begin{tabular}{C{0.8in} *4{C{.9in}}}\toprule[1.5pt]
\bf Tasks & \bf Packing  & \bf Planning & \bf Binary & \bf Max Cut \\\midrule
\bf Size & \bf $60 \times 60$  & \bf $122 \times 168$ & \bf $66 \times 132$ & \bf $54 \times 134$  \\\midrule
No cut   &  $0.26 \pm 0.09$  & $0.25\pm 0.04$ & $0.74 \pm 0.24$ & $0.95 \pm 0.09$ \\
Random  &  $0.31 \pm 0.08$  & $0.65\pm 0.10$ & $0.94 \pm 0.10$ & $0.99 \pm 0.04$  \\
MV & $0.23 \pm 0.08$ & $0.27 \pm 0.07$ & $0.92 \pm 0.12$ & $0.98 \pm 0.06$  \\ 
MNV & $0.27 \pm 0.08$ & $0.33 \pm 0.15$ & $0.93 \pm 0.10$ & $1.0 \pm 0.0$ \\ 
LE & $0.28 \pm 0.08$ & $0.25 \pm 0.04$ & $0.95 \pm 0.08$ & $0.95 \pm 0.09$ \\ 
RL & $\mathbf{0.36 \pm 0.10}$ & $\mathbf{0.99 \pm 0.02}$ & $\mathbf{0.96 \pm 0.08}$ & $1.0 \pm 0.0$ \\
\bottomrule
\end{tabular}
\end{sc}
\end{small}
\end{center}
\vskip -0.1in
\label{table:numberofnodes_largescale}
\end{table}

\section{Comparison of Distributed Agent Interface}
\label{appendix:agent}
To scale \gls{RL} training to powerful computational architecture, it is imperative that the agent becomes distributed. Indeed, recent years have witnessed an increasing attention on the design and implementation of distributed algorithms \citep{mnih2016asynchronous,salimans2017evolution,espeholt2018impala,kapturowski2018recurrent}.

 General distributed algorithms adopt a learner-actor architecture, i.e. one central learner and multiple distributed actors. Actors collect data and send partial trajectories to the learner. The learner takes data from all actors and generates updates to the central parameter. The general interface requires a function $\pi_\theta(a|s)$ parameterized by $\theta$, which takes a state $s$ and outputs an action $a$ (or a distribution over actions). In a gradient-based algorithm (e.g.~\citep{espeholt2018impala}), the actor executes such an interface with the forward mode and generates trajectory tuple $(s,a)$; the learner executes the interface with the backward mode to compute gradients and update $\theta$. Below we list several practical considerations why \gls{ES} is a potentially better distributed alternative to such gradient-based distributed algorithms in this specific context, where state/action spaces are irregular.

\begin{itemize}
      \item \textbf{Communication.} The data communication between learner-actor is more complex for general gradient-based algorithms. Indeed, actors need to send partial trajectories $\{ (s_i,a_i,r_i) \}_{i=1}^\tau$ to the learner, which requires careful adaptations to cases where the state/action space are irregular. On the other hand, \gls{ES} only require sending returns over trajectories $\sum_{i=0}^T r_i$, which greatly simplifies the interface from an engineering perspective.
      \item \textbf{Updates.} Gradient-based updates require both forward/backward mode of the agent interface. Further, the backward mode function needs to be updated such that batched processing is efficient to allow for fast updates. For irregular state/action space, this requires heavier engineering because of e.g.~arrays of variable sizes are not straightforward to be batched. On the other hand, \gls{ES} only requires forward mode computations required by CPU actors.
\end{itemize}  
  
\section{Considerations on CPU Runtime}

In practice, instead of the number of cuts, a more meaningful budget constraint on solvers is the CPU runtime, i.e. practitioners typically set a runtime constraint on the solver and expect the solver to return the best possible solution within this constraint. Below, we report runtime results for training/test time. We will show that even under runtime constraints, the \gls{RL} policy achieves significant performance gains. 

\paragraph{Training time.}
During training time, it is not straightforward to explicitly maintain a constraint on the runtime, because it is very sensitive to hardware conditions (e.g.~number of available processors). Indeed, prior works \citep{khalil2016learning,dai2017learning} do not apply runtime constraint during training time, though runtime constraint is an important measure at test time.

The absolute training time depends on specific hardware architecture. In our experiments we train with a single server with $64$ virtual CPUs. Recall that each update consists in collecting trajectories across training instances and generating one single gradient update. We observe that typically the convergence takes place in $\leq 500$ weight updates (iterations).

\paragraph{Test time.}
To account for the practical effect of runtime, we need to account for the following trade-off: though RL based policy produces higher-quality cutting planes in general, running the policy at test time could be costly. To characterize the trade-offs, we address the following question:  \textbf{(1)} When adding a fixed number of cuts, does RL lead to higher runtime? \textbf{(2)} When solving a particular problem, does RL lead to performance gains in terms of runtime?

To address \textbf{(1)}, we reuse the experiments in Experiment \#2, i.e. adding a fixed number of cuts $T=50$ on middle sized problems. The runtime results are presented in Table \ref{table:runtime_fixedcuts}, where we show that \gls{RL} cutting plane selection does not increase the runtime significantly compared to other 'fast' heuristics. Indeed, \gls{RL} increases the average runtime in some cases while decreases in others. Intuitively, we expect the runtime gains to come from the fact that RL requires a smaller number of cuts - leading to fewer iterations of the algorithm. However, this is rare in Experiment \#2, where for most instances optimal solution is not reached in maximum number of cuts, so all heuristics and RL add same number of cuts ($T=50$). We expect such advantages to become more significant with the increase of the size of the problem, as the computational gain of adding good cuts becomes more relevant. We confirm such intuitions from the following.

To address \textbf{(2)}, we reuse the results from Experiment \#4, where we solve more difficult instances with \gls{bnc}, we report the runtime results in Table \ref{table:runtime_bc}. In these cases, the benefits of high-quality cuts are magnified by a decreased number of iterations (i.e. expanded nodes) - indeed, for RL policy, the advantages resulting from decreased iterations significantly overweight the potentially slight drawbacks of per-iteration runtime. In Table \ref{table:runtime_bc}, we see that RL generally requires much smaller runtime than other heuristics, mainly due to a much smaller number of \gls{bnc} iterations. Note that these results are consistent with Figure~\ref{figure:numberofbb}. Again, for large-scale problems, this is an important advantage in terms of usage of memory and overall performance of the system.


\begin{table}[t]
\caption{CPU runtime  for adding cutting planes (units are seconds). Here we present the results from Experiment \#2 from the main text, where we fix the number of added cuts $T=50$. Note that though RL might increase runtime in certain cases, it achieves much larger IGC within the cut budgets. Note that these results are consistent with Table \ref{table:optimalitygap_medium}.}
\begin{center}
\begin{small}
\begin{sc}
\begin{tabular}{C{0.8in} *4{C{.9in}}}\toprule[1.5pt]
\bf Tasks & \bf Packing  & \bf Planning & \bf Binary & \bf Max Cut \\\midrule
\bf Size & \bf $30 \times 30$  & \bf $61 \times 84$ & \bf $33 \times 66$ & \bf $27 \times 67$  \\\midrule
Random   &  $0.06 \pm 0.01$  & $0.09 \pm 0.01$ & $0.088 \pm 0.003$ & $0.08 \pm 0.01$ \\
MV & $0.9 \pm 0.01$ & $0.100 \pm 0.004$ & $0.10 \pm 0.01$ & $0.11 \pm 0.01$  \\ 
MNV & $0.10 \pm 0.02$ & $0.100 \pm 0.004$ & $0.12\pm 0.02$ & $0.12 \pm 0.01$\% \\ 
RL & $0.10 \pm 0.02$ & $0.14 \pm 0.03$ & $0.07 \pm 0.04$ & $0.08 \pm 0.02$ \\
\bottomrule
\end{tabular}
\end{sc}
\end{small}
\end{center}
\vskip -0.1in
\label{table:runtime_fixedcuts}
\end{table}

\begin{table}[t]
\caption{CPU runtime in \gls{bnc} with large-scale instances. The measures are normalized with respect to RL so that the RL runtime is always  measured as $100$\%. Here, we measure the runtime as the time it takes to reach a certain level of IGC. We only measure the runtime on test instances where the IGC level is reached within the node budgets. When the IGC is not reached for most test instances (as in the case of the planning problem for most baselines), the runtime measure is 'N/A'. Note that the results here are consistent with Table \ref{table:optimalitygap_largescale} and Figure \ref{figure:numberofbb_largescale}.}
\begin{center}
\begin{small}
\begin{sc}
\begin{tabular}{C{0.8in} *4{C{.9in}}}\toprule[1.5pt]
\bf Tasks & \bf Packing  & \bf Planning & \bf Binary & \bf Max Cut \\\midrule
\bf Size & \bf $60 \times 60$  & \bf $122 \times 168$ & \bf $66 \times 132$ & \bf $54 \times 134$  \\\midrule
Random  &  $146\%$  & $\text{N/A}$ & $190 \%$ & $250$\%  \\
MV & $256\%$ & $\text{N/A}$ & $340 \%$ & $210$\%  \\ 
MNV & $238\%$ & $\text{N/A}$ & $370 \%$ & $\mathbf{95}$\% \\ 
LE & $120\%$ & $\text{N/A}$ & $370$\% & $120$\% \\ 
RL & $\mathbf{100\%}$ & $\mathbf{100 \%}$ & $\mathbf{100\%}$ & $100\%$ \\
\bottomrule
\end{tabular}
\end{sc}
\end{small}
\end{center}
\vskip -0.1in
\label{table:runtime_bc}
\end{table}

\end{document}